# From Bayesian Sparsity to Gated Recurrent Nets

**Hao He**
Peking University, Beijing, China
hehaodele@pku.edu.cn

**Bo Xin**
Microsoft Research, Beijing, China
jimxinbo@gmail.com

**David Wipf**
Microsoft Research, Beijing, China
davidwipf@gmail.com

## Abstract

The iterations of many first-order algorithms, when applied to minimizing common regularized regression functions, often resemble neural network layers with pre-specified weights. This observation has prompted the development of learning-based approaches that purport to replace these iterations with enhanced surrogates forged as DNN models from available training data. For example, important NP-hard sparse estimation problems have recently benefitted from this genre of upgrade, with simple feedforward or recurrent networks ousting proximal gradient-based iterations. Analogously, this paper demonstrates that more powerful Bayesian algorithms for promoting sparsity, which rely on complex multi-loop majorization-minimization techniques, mirror the structure of more sophisticated long short-term memory (LSTM) networks, or alternative gated feedback networks previously designed for sequence prediction. As part of this development, we examine the parallels between latent variable trajectories operating across multiple time-scales during optimization, and the activations within deep network structures designed to adaptively model such characteristic sequences. The resulting insights lead to a novel sparse estimation system that, when granted training data, can estimate optimal solutions efficiently in regimes where other algorithms fail, including practical direction-of-arrival (DOA) and 3D geometry recovery problems. The underlying principles we expose are also suggestive of a learning process for a richer class of multi-loop algorithms in other domains.

## 1 Introduction

Many practical iterative algorithms for minimizing an energy function $\mathcal{L}_y(\boldsymbol{x})$, parameterized by some vector $\boldsymbol{y}$, adopt the updating prescription

$$\boldsymbol{x}^{(t+1)} = f(\boldsymbol{A}\boldsymbol{x}^{(t)} + \boldsymbol{B}\boldsymbol{y}), \tag{1}$$

where $t$ is the iteration count, $\boldsymbol{A}$ and $\boldsymbol{B}$ are fixed matrices/filters, and $f$ is a point-wise nonlinear operator. When we treat $\boldsymbol{B}\boldsymbol{y}$ as a bias or exogenous input, then the progression of these iterations through time resembles activations passing through the layers (indexed by $t$) of a deep neural network (DNN) [23, 35, 39, 2]. It then naturally begs the question: If we have access to an ensemble of pairs $\{\boldsymbol{y}, \boldsymbol{x}^*\}$, where $\boldsymbol{x}^* = \arg\min_{\boldsymbol{x}} \mathcal{L}_y(\boldsymbol{x})$, can we train an appropriately structured DNN to produce a minimum of $\mathcal{L}_y(\boldsymbol{x})$ when presented with an arbitrary new $\boldsymbol{y}$ as input? If $\boldsymbol{A}$ and $\boldsymbol{B}$ are fixed for all $t$, this process can be interpreted as training a recurrent neural network (RNN), while if they vary, a deep feedforward network with independent weights on each layer is a more apt description.

Although many of our conclusions may ultimately have broader implications, in this work we focus on minimizing the ubiquitous sparse estimation problem

$$\mathcal{L}_y(\boldsymbol{x}) = \|\boldsymbol{y} - \boldsymbol{\Phi}\boldsymbol{x}\|_2^2 + \lambda\|\boldsymbol{x}\|_0, \tag{2}$$

where $\boldsymbol{\Phi} \in \mathbb{R}^{n\times m}$ is an overcomplete matrix of feature vectors, $\|\cdot\|_0$ is the $\ell_0$ norm equal to a count of the nonzero elements in a vector, and $\lambda > 0$ is a trade-off parameter. Although crucial to many

applications [3, 12, 16, 20, 26, 31], solving (2) is NP-hard, and therefore efficient approximations are sought. Popular examples with varying degrees of computational overhead include convex relaxations such as $\ell_1$-norm regularization [5, 11, 37] and many flavors of iterative hard-thresholding (IHT) [6, 7].

In most cases, these approximate algorithms can be implemented via (1), where $\boldsymbol{A}$ and $\boldsymbol{B}$ are functions of $\boldsymbol{\Phi}$, and the nonlinearity $f$ is, for example, a hard-thresholding operator for IHT or soft-thresholding for convex relaxations. However, the Achilles' heel of all these approaches is that they will generally not converge to good approximate minimizers of (2) if $\boldsymbol{\Phi}$ has columns with a high degree of correlation [6, 11], which is unfortunately often the case in practice [40].

To mitigate the effects of such correlations, we could leverage the aforementioned correspondence with common DNN structures to learn something like a correlation-invariant algorithm or update rules [2], although in this scenario our starting point would be an algorithmic format with known deficiencies. But if our ultimate goal is to learn a new sparse estimation algorithm that efficiently compensates for structure in $\boldsymbol{\Phi}$, then it seems reasonable to invoke iterative algorithms known *a priori* to handle such correlations directly as our template for learned network layers. One important example is sparse Bayesian learning (SBL) [38], which has been shown to solve (2) using a principled, multi-loop majorization-minimization approach [25] even in cases where $\boldsymbol{\Phi}$ displays strong correlations [40].[1] Herein we demonstrate that, when judiciously unfolded, SBL iterations can be formed into variants of long short-term memory (LSTM) cells, one of the more popular recurrent deep neural network architectures [24], or gated extensions thereof [15]. The resulting network dramatically outperforms existing methods in solving (2) with a minimal computational budget. Our high-level contributions can be summarized as follows:

- Quite surprisingly, we demonstrate that the SBL objective, which explicitly compensates for correlated dictionaries, can be optimized using iteration structures that map directly to popular LSTM cells despite its radically different origin. This association significantly broadens recent work associating elementary, one-step iterative sparsity algorithms like (1) with simple recurrent or feedforward deep network architectures [23, 35, 39, 2].
- At its core, any SBL algorithm requires coordinating inner- and outer-loop computations that produce expensive latent posterior variances (or related, derived quantities) and optimized coefficient estimates respectively. Although this process can in principle be accommodated via canonical LSTM cells, such an implementation will enforce that computation of latent variables rigidly map to predefined subnetworks corresponding with various gating structures, ultimately administering a fixed schedule of switching between loops. To provide greater flexibility in coordinating inner- and outer-loops, we propose a richer gated-feedback LSTM structure for sparse estimation.
- We achieve state-of-the-art performance on several empirical tasks, including direction-of-arrival (DOA) estimation [32] and 3D geometry recovery via photometric stereo [43]. In these and other cases, our approach produces higher accuracy estimates at a fraction of the computational budget. These results are facilitated by a novel online data generation process.
- Although learning-to-learn style approaches [1, 23, 35, 39] have been commonly applied to relatively simple gradient descent optimization templates, this is the first successful attempt we are aware of to learn a complex, multi-loop, majorization-minimization algorithm [25]. We envision that such a strategy can have wide-ranging implications beyond the sparse estimation problems explored herein given that it is often not obvious how to optimally tune loop execution to balance both complexity and estimation accuracy in practice.

## 2 Connecting SBL and LSTM Networks

This section first reviews the basic SBL model, followed an algorithmic characterization of how correlation structure can be handled during sparse estimation. Later we derive specialized SBL update rules that reveal a close association with LSTM cells.

[1]Note also that a recent interesting modification of approximate message passing [34] (or its unfolded, trainable deep analog that converges to the same solution [9]), can handle certain specialized forms of dictionary correlation; however, the approach does not work with the types of strong arbitrary/unconstrained correlation we consider in this work.

## 2.1 Original SBL Model

Given an observed vector $\boldsymbol{y} \in \mathbb{R}^n$ and feature dictionary $\boldsymbol{\Phi} \in \mathbb{R}^{n\times m}$, SBL assumes the Gaussian likelihood model and a parameterized zero-mean Gaussian prior for the unknown coefficients $\boldsymbol{x} \in \mathbb{R}^m$ given by

$$p(\boldsymbol{y}|\boldsymbol{x}) \propto \exp\left[-\tfrac{1}{2\lambda}\|\boldsymbol{y}-\boldsymbol{\Phi x}\|_2^2\right] \quad \text{and} \quad p(\boldsymbol{x};\boldsymbol{\gamma}) \propto \exp\left[-\tfrac{1}{2}\boldsymbol{x}^\top \boldsymbol{\Gamma}^{-1}\boldsymbol{x}\right], \;\; \boldsymbol{\Gamma} \triangleq \text{diag}[\gamma], \tag{3}$$

where $\lambda > 0$ is a fixed variance factor and $\boldsymbol{\gamma}$ denotes a vector of unknown hyperparamters [38]. Because both likelihood and prior are Gaussian, the posterior $p(\boldsymbol{x}|\boldsymbol{y};\boldsymbol{\gamma})$ is also Gaussian, with mean $\hat{\boldsymbol{x}}$ satisfying

$$\hat{\boldsymbol{x}} = \boldsymbol{\Gamma\Phi}^\top \boldsymbol{\Sigma}_y^{-1}\boldsymbol{y}, \quad \text{with } \boldsymbol{\Sigma}_y \triangleq \boldsymbol{\Phi\Gamma\Phi}^\top + \lambda \boldsymbol{I}. \tag{4}$$

Given the lefthand-side multiplication by $\boldsymbol{\Gamma}$ in (4), $\hat{\boldsymbol{x}}$ will have a matching sparsity profile or support pattern as $\boldsymbol{\gamma}$, meaning that the locations of zero-valued elements will align or supp$[\hat{\boldsymbol{x}}]$ = supp$[\boldsymbol{\gamma}]$. Ultimately then, the SBL strategy shifts from directly searching for some optimally sparse $\hat{\boldsymbol{x}}$, to an optimally sparse $\boldsymbol{\gamma}$. For this purpose we marginalize over $\boldsymbol{x}$ and then maximize the resulting type-II likelihood function with respect to $\boldsymbol{\gamma}$ [30]. Conveniently, the resulting convolution-of-Gaussians integral is available in closed-form [38] such that we can equivalently minimize the negative log-likelihood

$$\mathcal{L}(\boldsymbol{\gamma}) \;=\; -\log \int p(\boldsymbol{y}|\boldsymbol{x})p(\boldsymbol{x};\boldsymbol{\gamma})d\boldsymbol{x} \;\equiv\; \boldsymbol{y}^\top \boldsymbol{\Sigma}_y^{-1}\boldsymbol{y} + \log|\boldsymbol{\Sigma}_y|. \tag{5}$$

Given an optimal $\boldsymbol{\gamma}$ so obtained, we can compute the posterior mean estimator $\hat{\boldsymbol{x}}$ via (4). Equivalently, this same posterior mean estimator can be obtained by an iterative reweighted $\ell_1$ process described next that exposes subtle yet potent sparsity-promotion mechanisms.

## 2.2 Iterative Reweighted $\ell_1$ Implementation

Although not originally derived this way, SBL can be implemented using a modified form of iterative reweighted $\ell_1$-norm optimization that exposes its agency for producing sparse estimates. In general, if we replace the $\ell_0$ norm from (2) with any smooth approximation $g(|\boldsymbol{x}|)$, where $g$ is a concave, non-decreasing function and $|\cdot|$ applies elementwise, then cost function descent[2] can be guaranteed using iterations of the form [41]

$$\boldsymbol{x}^{(t+1)} \leftarrow \arg\min_{\boldsymbol{x}} \|\boldsymbol{y}-\boldsymbol{\Phi x}\|_2^2 + \lambda \textstyle\sum_i w_i^{(t)}|x_i|, \qquad w_i^{(t+1)} \leftarrow \partial g(\boldsymbol{u})/\partial u_i\big|_{u_i = \left|x_i^{(t+1)}\right|}, \quad \forall i. \tag{6}$$

This process can be viewed as a multi-loop, majorization-minimization algorithm [25] (a generalization of the EM algorithm [18]), whereby the inner-loop involves computing $\boldsymbol{x}^{(t+1)}$ by minimizing a first-order, upper-bounding approximation $\|\boldsymbol{y}-\boldsymbol{\Phi x}\|_2^2 + \lambda\sum_i w_i^{(t)}|x_i|$, while the outer-loop updates the bound/majorizer itself as parameterized by the weights $\boldsymbol{w}^{(t+1)}$. Obviously, if $g(\boldsymbol{u}) = \boldsymbol{u}$, then $\boldsymbol{w}^{(t)} = \mathbf{1}$ for all $t$, and (6) reduces to the Lasso objective for $\ell_1$ norm regularized sparse regression [37], and only a single iteration is required. However, one popular non-trivial instantiation of this approach assumes $g(\boldsymbol{u}) = \sum_i \log(u_i + \epsilon)$ with $\epsilon > 0$ a user-defined parameter [13]. The corresponding weights then become $w_i^{(t+1)} = \left(\left|x_i^{(t+1)}\right| + \epsilon\right)^{-1}$, and we observe that once any particular $x_i^{(t+1)}$ becomes large, the corresponding weight becomes small and at the next iteration a weaker penalty will be applied. This prevents the overshrinkage of large coefficients, a well-known criticism of $\ell_1$ norm penalties [19].

In the context of SBL, there is no closed-form $w_i^{(t+1)}$ update except in special cases. However, if we allow for additional latent structure, which we later show is akin to the memory unit of LSTM cells, a viable recurrency emerges for computing these weights and elucidating their effectiveness in dealing with correlated dictionaries. In particular we have:

**Proposition 1.** *If weights* $\boldsymbol{w}^{(t+1)}$ *satisfy*

$$\left(w_i^{(t+1)}\right)^2 = \min_{\boldsymbol{z}:supp[\boldsymbol{z}]\subseteq supp[\boldsymbol{\gamma}^{(t)}]} \tfrac{1}{\lambda}\|\boldsymbol{\phi}_i - \boldsymbol{\Phi z}\|_2^2 + \textstyle\sum_{j\in supp[\boldsymbol{\gamma}^{(t)}]} \frac{z_j^2}{\gamma_j^{(t+1)}} \tag{7}$$

[2] Or global convergence to some stationary point with mild additional assumptions [36].

*for all i, then the iterations (6), with* $\gamma_j^{(t+1)} = \left[w_j^{(t)}\right]^{-1/2} \left|x_j^{(t+1)}\right|$*, are guaranteed to reduce or leave unchanged the SBL objective (5). Also, at each iteration,* $\boldsymbol{\gamma}^{(t+1)}$ *and* $\boldsymbol{x}^{(t+1)}$ *will satisfy (4).*

Unlike the traditional sparsity penalty mentioned above, with SBL we see that the $i$-th weight $w_i^{(t+1)}$ is not dependent solely on the value of the $i$-th coefficient $x_i^{(t+1)}$, but rather on *all* the latent hyperparameters $\boldsymbol{\gamma}^{(t+1)}$ and therefore ultimately prior-iteration weights $\boldsymbol{w}^{(t)}$ as well. Moreover, because the fate of each sparse coefficient is linked together, correlation structure can be properly accounted for in a progressive fashion.

More concretely, from (7) it is immediately apparent that if $\boldsymbol{\phi}_i \approx \boldsymbol{\phi}_{i'}$ for some indeces $i$ and $i'$ (meaning a large degree of correlation), then it is highly likely that $w_i^{(t+1)} \approx w_{i'}^{(t+1)}$. This is simply because the regularized residual error that emerges from solving (7) will tend to be quite similar when $\boldsymbol{\phi}_i \approx \boldsymbol{\phi}_{i'}$. In this situation, a suboptimal solution will not be prematurely enforced by weights with large, spurious variance across a correlated group of basis vectors. Instead, weights will differ substantially only when the corresponding columns have meaningful differences relative to the dictionary as a whole, in which case such differences can help to avoid overshrinkage as before.

A crucial exception to this perspective occurs when $\boldsymbol{\gamma}^{(t+1)}$ is highly sparse, or nearly so, in which case there are limited degrees of freedom with which to model even small differences in each $\boldsymbol{\phi}_i$. However, such cases can generally only occur when we are in the neighborhood of ideal, maximally sparse solutions by definition [40], when different weights are actually desirable even among correlated columns for resolving the final sparse estimates.

### 2.3 Revised SBL Iterations

The iterative reweighted $\ell_1$ formulation of SBL, as elucidated as we have done through (7), can no longer be implemented via the simple recurrent structures used in the past, e.g., like (1) for learning optimal IHT or other sparsity promoting iterations. Instead, the additional latent dependencies that arise through $\boldsymbol{w}$, $\boldsymbol{x}$, and $\boldsymbol{\gamma}$, which allow SBL to incrementally store or update learned representations (akin to incrementally accruing an optimal sparsity profile), require a more sophisticated network architecture to actualize.

Although presumably there are multiple ways such an architecture could be developed, in this section we derive specialized SBL iterations that will directly map to one of the most common RNN structures, namely LSTM networks. With this in mind, the notation we adopt has been intentionally chosen to facilitate later association with LSTM cells. We first define

$$\boldsymbol{w}^{(t)} \triangleq \text{diag}\left[\boldsymbol{\Phi}^\top \left(\lambda \boldsymbol{I} + \boldsymbol{\Phi}\boldsymbol{\Gamma}^{(t)}\boldsymbol{\Phi}^\top\right)^{-1} \boldsymbol{\Phi}\right]^{\frac{1}{2}} \quad \text{and} \quad \boldsymbol{\nu}^{(t)} \triangleq \boldsymbol{u}^{(t)} + \mu \boldsymbol{\Phi}^\top \left(\boldsymbol{y} - \boldsymbol{\Phi}\boldsymbol{u}^{(t)}\right), \tag{8}$$

where $\boldsymbol{\Gamma}^{(t)} \triangleq \text{diag}\left[\boldsymbol{\gamma}^{(t)}\right]$, $\boldsymbol{u}^{(t)} \triangleq \boldsymbol{\Gamma}^{(t)}\boldsymbol{\Phi}^\top \left(\lambda \boldsymbol{I} + \boldsymbol{\Phi}\boldsymbol{\Gamma}^{(t)}\boldsymbol{\Phi}^\top\right)^{-1} \boldsymbol{y}$, and $\mu > 0$ is a constant. As will be discussed further below, $\boldsymbol{w}^{(t)}$ serves the exact same role as the weights from (7), hence the identical notation. We then partition our revised SBL iterations as so-called *gate* updates

$$\boldsymbol{\sigma}_{in}^{(t+1)} \leftarrow \left[\boldsymbol{\alpha}\left(\boldsymbol{\gamma}^{(t)}\right) \odot \left(\left|\boldsymbol{\nu}^{(t)}\right| - 2\lambda \boldsymbol{w}^{(t)}\right)\right]_+, \qquad \boldsymbol{\sigma}_{f}^{(t+1)} \leftarrow \boldsymbol{\beta}\left(\boldsymbol{\gamma}^{(t)}\right), \qquad \boldsymbol{\sigma}_{out}^{(t+1)} \leftarrow \left(\boldsymbol{w}^{(t)}\right)^{-1}, \tag{9}$$

*cell* updates

$$\bar{\boldsymbol{x}}^{(t+1)} \leftarrow \text{sign}\left[\boldsymbol{\nu}^{(t)}\right], \qquad \boldsymbol{x}^{(t+1)} \leftarrow \boldsymbol{\sigma}_f^{(t)} \odot \boldsymbol{x}^{(t)} + \boldsymbol{\sigma}_{in}^{(t)} \odot \bar{\boldsymbol{x}}^{(t+1)}, \tag{10}$$

and *output* update

$$\boldsymbol{\gamma}^{(t+1)} \quad \leftarrow \quad \boldsymbol{\sigma}_{out}^{(t)} \odot \left|\boldsymbol{x}^{(t+1)}\right|, \tag{11}$$

where the inverse and absolute-value operators are applied element-wise when a vector is the argument, and at least for now, $\boldsymbol{\alpha}$ and $\boldsymbol{\beta}$ define arbitrary functions. Moreover, $\odot$ denotes the Hadamard product and $[\cdot]_+$ sets negative values to zero and leaves positive quantities unchanged, also in an element-wise fashion, i.e., it acts just like a rectilinear (ReLU) unit [33]. Note also that the gate and cell updates in isolation can be viewed as computing a first-order, partial solution to the inner-loop weighted $\ell_1$ optimization problem from (6).

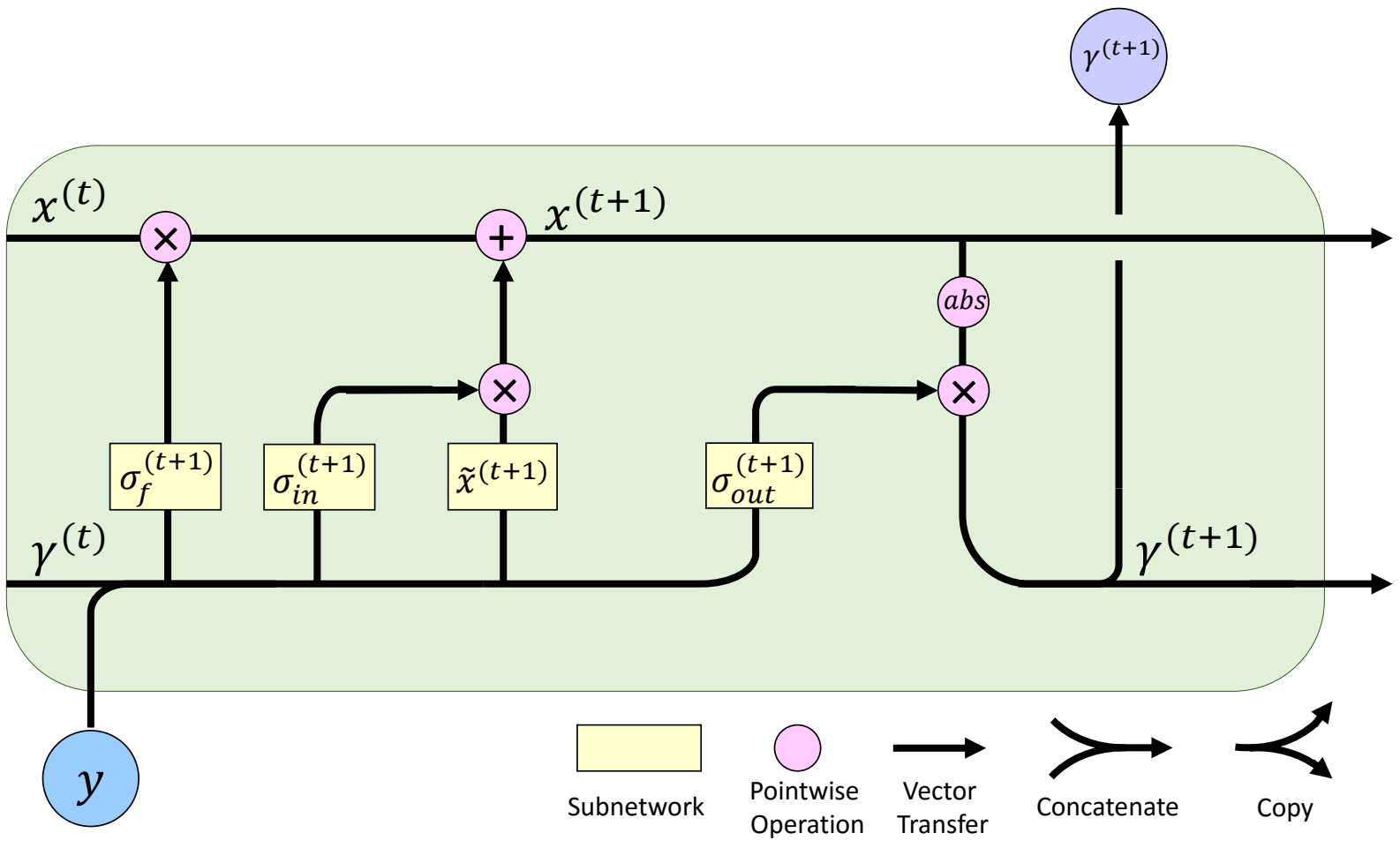

Figure 1: LSTM/SBL Network

Starting from some initial $\boldsymbol{\gamma}^{(0)}$ and $\boldsymbol{x}^{(0)}$, we will demonstrate in the next section that these computations closely mirror a canonical LSTM network unfolded in time with $\boldsymbol{y}$ acting as a constant input applied at each step. Before doing so however, we must first demonstrate that (8)−(11) indeed serve to reduce the SBL objective. For this purpose we require the following definition:

**Definition 2.** *We say that the iterations (9)−(11) satisfy the monotone cell update property if*

$$\|\boldsymbol{y} - \boldsymbol{\Phi}\boldsymbol{u}^{(t)}\|_2^2 + 2\lambda \sum_i w_i^{(t)} |u_i^{(t)}| \;\; \geq \;\; \|\boldsymbol{y} - \boldsymbol{\Phi}\boldsymbol{x}^{(t+1)}\|_2^2 + 2\lambda \sum_i w_i^{(t)} |x_i^{(t+1)}|, \;\; \forall t. \tag{12}$$

Note that for rather inconsequential technical reasons this definition involves $\boldsymbol{u}^{(t)}$, which can be viewed as a proxy for $\boldsymbol{x}^{(t)}$. We then have the following:

**Proposition 3.** *The iterations (8)−(11) will reduce or leave unchanged (5) for all $t$ provided that $\mu \in \left(0, \lambda / \left\| \boldsymbol{\Phi}^\top \boldsymbol{\Phi} \right\|\right]$ and $\boldsymbol{\alpha}$ and $\boldsymbol{\beta}$ are chosen such that the monotone cell update property holds.*

In practical terms, the simple selections $\boldsymbol{\alpha}(\boldsymbol{\gamma}) = \mathbf{1}$ and $\boldsymbol{\beta}(\boldsymbol{\gamma}) = \mathbf{0}$ will provably satisfy the monotone cell update property (see proof details in Appendix E). However, for additional flexibility, $\boldsymbol{\alpha}$ and $\boldsymbol{\beta}$ could be selected to implement various forms of momentum, ultimately leading to cell updates akin to the popular FISTA [5] or monotonic FISTA [4] algorithms. In both cases, old values $\boldsymbol{x}^{(t)}$ are precisely mixed with new factors $\boldsymbol{\sigma}_{in}^{(t+1)} \odot \bar{\boldsymbol{x}}^{(t+1)}$ to speed convergence. Of course the whole point of casting the SBL iterations as an RNN structure to begin with is so that we may ultimately *learn* these types of functions, without the need for hand-crafting suboptimal iterations up front.

### 2.4 Correspondences with LSTM Components

This section will flesh out how the SBL iterations presented in Section 2.3 display the same structure as a canonical LSTM cell, the only differences being the shape of the nonlinearities, and the exact details of the gate subnetworks. To facilitate this objective, Figure 1 contains a canonical LSTM network structure annotated with SBL-derived quantities. We now walk through these correspondences.

First, the exogenous input to the network is the observation vector $\boldsymbol{y}$, which does not change from time-step to time-step. This is much like the strategy used by feedback networks for obtaining incrementally refined representations [46]. The output at time-step $t$ is $\boldsymbol{\gamma}^{(t)}$, which serves as the current estimate of the SBL hyperparameters. In contrast, we treat $\boldsymbol{x}^{(t)}$ as the internal LSTM memory cell, or the latent cell state.[3] This deference to $\boldsymbol{\gamma}^{(t)}$ mirrors the emphasis SBL places on learning variances while treating $\boldsymbol{x}$ as hidden data, and in some sense flips the coefficient-centric script used in producing (6).

[3] If we allow for peephole connections [21], it is possible to reverse these roles; however, for simplicity and the most direct mapping to LSTM cells we do not pursue this alternative here.

Proceeding further, $\boldsymbol{\gamma}^{(t)}$ is fed to four separate layers/subnetworks (represented by yellow boxes in Figure 1): (i) the *forget* gate $\boldsymbol{\sigma}_f$, (ii) the *input* gate $\boldsymbol{\sigma}_{in}$, (iii) the *output* gate $\boldsymbol{\sigma}_{out}$, and (iv) the candidate input update $\bar{\boldsymbol{x}}$. The forget gate computes scaling factors for each element of $\boldsymbol{x}^{(t)}$, with small values of the gate output suggesting that we 'forget' the corresponding old cell state elements. Similarly the input gate determines how large we rescale signals from the candidate input update. These two re-weighted quantities are then mixed together to form the new cell state $\boldsymbol{x}^{(t+1)}$. Finally, the output gate modulates how new $\boldsymbol{\gamma}^{(t+1)}$ are created as scaled versions of the updated cell state.

Regarding details of these four subnetworks, based on the update templates from (9) and (10), we immediately observe that the required quantities depend directly on (8). Fortunately, both $\boldsymbol{\nu}^{(t)}$ and $\boldsymbol{w}^{(t)}$ can be naturally computed using simple feedforward subnetwork structures.[4] These values can either be computed in full (ideal case), or partially to reduce the computational burden. In any event, once obtained, the respective gates and candidate cell input updates can be computed by applying final non-linearities. Note that $\boldsymbol{\alpha}$ and $\boldsymbol{\beta}$ are treated as arbitrary subnetwork structures at this point that can be learned.

A few cosmetic differences remain between this SBL implementation and a canonical LSTM network. First, the final non-linearity for LSTM gating subnetworks is often a sigmoidal activation, whereas SBL is flexible with the forget gate (via $\boldsymbol{\beta}$), while effectively using a ReLU unit for the input gate and an inverse function for the output gate. Moreover, for the candidate cell update subnetwork, SBL replaces the typical tanh nonlinearity with a quantized version, the sign function, and likewise, for the output nonlinearity an absolute value operator (abs) is used. Finally, in terms of internal subnetwork structure, there is some parameter sharing given that $\boldsymbol{\sigma}_{in}$, $\boldsymbol{\sigma}_{out}$, and $\bar{\boldsymbol{x}}$ are connected via $\boldsymbol{\nu}$ and $\boldsymbol{w}$.

Of course in all cases we need not necessarily share parameters nor abide by these exact structures. In fact there is nothing inherently optimal about the particular choices used by SBL; rather it is merely that these structures happen to reproduce the successful, yet hand-crafted SBL iterations. But certainly there is potential in replacing such iterations with learned LSTM-like surrogates, at least when provided with access to sufficient training data as in prior attempts to learn sparse estimation algorithms [23, 39, 2].

## 3 The Dynamics of SBL Iterations

Although SBL iterations can be molded into an LSTM structure as we have shown, there remain hints that the full potential of this association may be presently undercooked. Here we empirically examine the trajectories of SBL iterations produced via the rules derived in Section 2.3. This process will unmask certain characteristic dynamics suggestive of a richer class of recurrent network structures, inspired by sequence prediction tasks [15], to be analyzed later in Section 4 and empirically tested in Section 5.

### 3.1 Large Timescale Differences

We first present a synthetic experiment that highlights the different time scales upon which SBL latent variables may fluctuate over the course of a typical optimization trajectory. The experimental design is as follows: First we generate a dictionary $\boldsymbol{\Phi}$ via

$$\boldsymbol{\Phi} = \widetilde{\boldsymbol{\Phi}} \boldsymbol{B} \boldsymbol{D}, \tag{13}$$

where $\widetilde{\boldsymbol{\Phi}} \in \mathbb{R}^{50 \times 100}$ has iid elements drawn from $\mathcal{N}(0, 1)$; $\boldsymbol{B} \in \mathbb{R}^{100 \times 100}$ is a block-diagonal matrix with 20, $5 \times 5$ blocks, each with unit diagonals and off-diagonals set to 0.9; and $\boldsymbol{D} \in \mathbb{R}^{100 \times 100}$ is a fully diagonal matrix that re-scales each column of the final $\boldsymbol{\Phi}$ to have unit $\ell_2$ norm, and finally multiplies by a random sign pattern. This process ensures that $\boldsymbol{\Phi}$ will encompass 20 clusters of 5 adjacent columns each, with strong correlations introduced via $\boldsymbol{B}$. We then generate a sparse random vector $\boldsymbol{x}^* \in \mathbb{R}^{100}$ such that $\|\boldsymbol{x}^*\|_0 = 10$, where the nonzero positions are randomly aligned with 10 different clusters, and the nonzero values have unit magnitude. We next compute $\boldsymbol{y} = \boldsymbol{\Phi}\boldsymbol{x}^*$ and apply the revised SBL iterations from Section 2.3 with $\lambda = 0.01$ (or a rather arbitrary small value), $\boldsymbol{\alpha}(\boldsymbol{\gamma}) = \boldsymbol{1}$, and $\boldsymbol{\beta}(\boldsymbol{\gamma}) = \boldsymbol{0}$.

[4] For $\boldsymbol{w}^{(t)}$ the result of Proposition 1 suggests that these weights can be computed as the solution of a simple regularized regression problem, which can easily be replaced with a small network analogous to that used in [21]; similarly for $\boldsymbol{\nu}$.

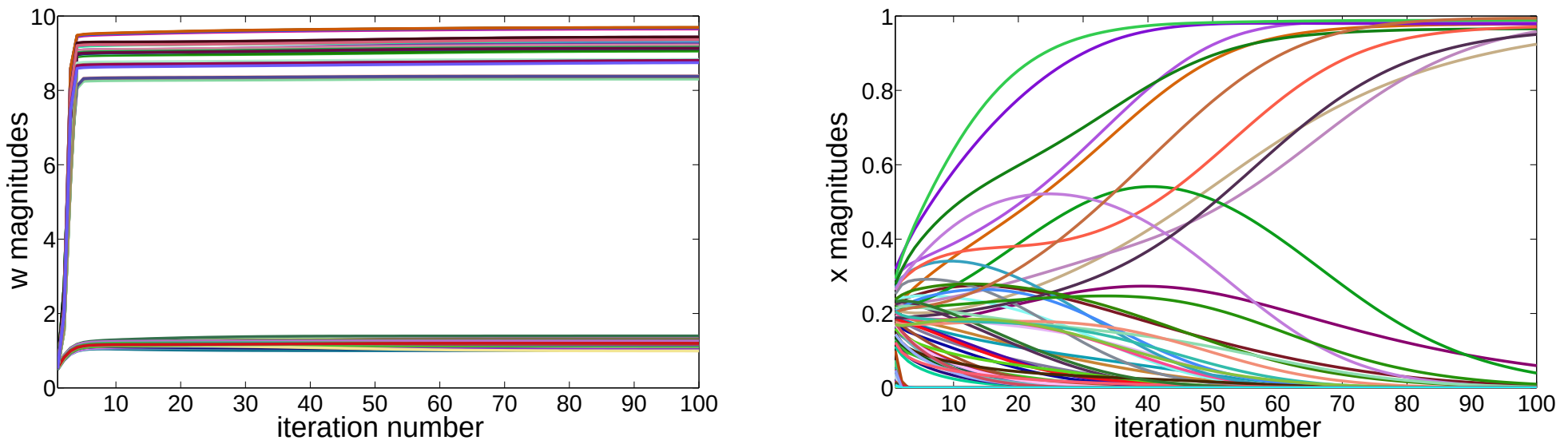

Figure 2: Illustration of the different time scales upon which SBL latent variables operate under a multi-resolution, clustered dictionary model. *Left*: Trajectories of the weights $\boldsymbol{w}$ across 100 iterations (each colored line represents a different element $w_i$). Within just a few iterations, these values have completely converged and accurately reflect the true cluster-level support pattern, namely, high values in the 8-10 range represent weights associated with false clusters (hence a high penalty/weight), while low values around 1.0 indicate weights associated with correct clusters. *Right*: Corresponding trajectories of $|\boldsymbol{x}|$ (a different colored line indicates a different $|x_i|$). Here we observe that even after 50 iterations, it is not entirely clear to what value each element will finally converge to. From these plots it is readily apparent that after 5 iterations, it is no longer necessary to update $\boldsymbol{w}$, provided the network is capable of memorizing the stored value from prior iterations, while $\boldsymbol{x}$ must be updated even beyond 100 iterations for full convergence.

Figure 2 displays the trajectories of both $\boldsymbol{w}^{(t)}$ (left subplot) and $\boldsymbol{x}^{(t)}$ (right subplot) for $t = 1, \ldots, 100$ during execution of (8)$-$(11). As mentioned previously, the weights $\boldsymbol{w}^{(t)}$ serve to incrementally focus a sequence of $\ell_1$ minimization problems towards likely nonzero elements of $\boldsymbol{x}^*$ via the process defined by (7). Unlike other existing iterative reweighted $\ell_1$ approaches, with SBL these weights quickly (within just a few iterations) partition into two groups, one with smaller values near 1.0, the other with larger values in the 8-10 range (see left subplot).

Moreover, upon closer examination we found that the index $i$ of all weights $w_i^{(t)}$ with a value near 1.0 correspond with dictionary columns $\boldsymbol{\phi}_i$ in a cluster where some $x_j^* \neq 0$. In contrast, all weights with a large value are associated with dictionary columns in clusters where all $x_j^* = 0$. Consequently, these weights reflect in some sense the correct support at the cluster level, and introduce a more severe penalty to coefficients associated with what should ideally be inactive clusters. This then allows subsequent $\ell_1$ iterations to more narrowly learn the correct support pattern *within* these favored clusters, providing empirical support to the arguments made in Section 2 for the efficacy of SBL in dealing with correlated dictionary structure.

However, the secondary learning of final coefficient values within the correct clusters occurs at a radically different time scale as shown in the right subplot of Figure 2. Here we observe that even after 50 iterations it is still not clear to which final value the coefficient magnitudes $x_i^{(t)}$ will converge too, for example, 0.0 or 1.0. Therefore we may conclude that, although the weights $\boldsymbol{w}^{(t)}$ may rather quickly proceed to values that reflect the correlation structure of the dictionary, the final coefficient estimates take much longer to resolve. Moreover, during this time, to be effective the iterations must 'remember' the correct value of $\boldsymbol{w}^{(t)}$, even if continued updates are not necessary after rapid initial convergence.

### 3.2 The Potential Value of an Adaptive Updating Schedule

Although the previous experiment served to expose the differing scales of subsets of latent variables, it did not provide any indication of how these different scales may actually impact final estimation accuracy. For example, suppose we were able to speed up the convergence of $\boldsymbol{x}$ for any fixed value of $\boldsymbol{w}$, would this improve the overall performance? The present simulation directly addresses this issue.

We begin with a similar experimental design as used in Section 3.1, although we reduce the dimensions for visualization purposes. In brief, we choose $\boldsymbol{\Phi} \in \mathbb{R}^{10\times 20}$, formed from 10 clusters of size 2 each, and $\|\boldsymbol{x}^*\|_0 = 3$. Figure 3(a) displays the optimal support pattern, whereby a '1' indicates the location of a true nonzero, and a zero otherwise. Without loss of generality, we have also reordered the dictionary columns such that the first three columns serve as nonzero locations.

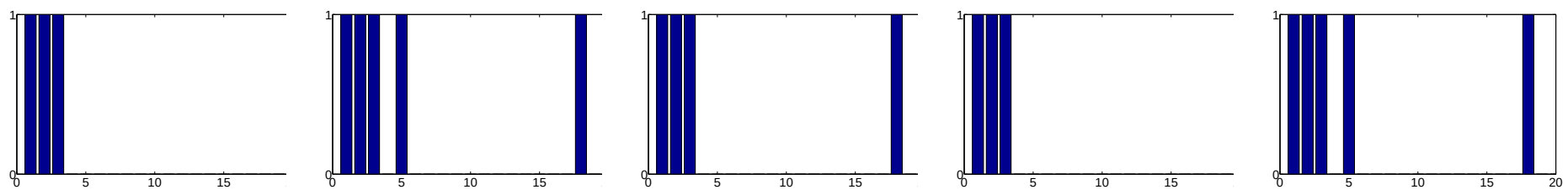

(a) Ground truth (b) $K = 1$ (c) $K = 10$ (d) $K = 100$ (e) $K = 1000$

Figure 3: Estimated support patterns obtained under varying numbers of inner-loop iterations $K$. A blue bar indicates that the corresponding index is associated with a nonzero element of $\hat{\boldsymbol{x}}$. We observe that when $K \in \{1, 10, 100, 1000\}$, the estimated support pattern is incorrect; see plots (b), (c), and (e). In contrast, $K = 100$ produces the correct result; see plot (d). Hence adaptively terminating these inner-loop iterations in a data-dependent fashion can potentially improve the final result.

We display recovery results using a modified version of the SBL implementation from Section 2, whereby the gate and cell update steps, which are associated with the weighted $\ell_1$ norm problem from (6), are applied in varying number $K$ of inner-loop iterations. More specifically, with $\boldsymbol{\alpha}(\boldsymbol{\gamma}) = \mathbf{1}$, and $\boldsymbol{\beta}(\boldsymbol{\gamma}) = \mathbf{0}$ and $\boldsymbol{w}^{(t)}$ fixed, these additional, reduced inner-loop iterations consist of simply computing

$$\begin{aligned}
\boldsymbol{\nu}^{(t,k)} &\leftarrow \boldsymbol{x}^{(t+1,k)} + \mu \boldsymbol{\Phi}^\top \left( \boldsymbol{y} - \boldsymbol{\Phi} \boldsymbol{x}^{(t+1,k)} \right) \\
\boldsymbol{\sigma}_{in}^{(t,k)} &\leftarrow \left[ \left| \boldsymbol{\nu}^{(t,k)} \right| - 2\lambda \boldsymbol{w}^{(t)} \right]_+ \\
\bar{\boldsymbol{x}}^{(t+1,k)} &\leftarrow \text{sign} \left[ \boldsymbol{\nu}^{(t,k)} \right] \\
\boldsymbol{x}^{(t+1,k+1)} &\leftarrow \boldsymbol{\sigma}_{in}^{(t,k)} \odot \bar{\boldsymbol{x}}^{(t+1,k)},
\end{aligned} \tag{14}$$

from $k = 1, \ldots, K$, where $\boldsymbol{x}^{(t+1,1)} \triangleq \boldsymbol{u}^{(t)}$. For any fixed $\boldsymbol{w}^{(t)}$, these iterations are guaranteed to converge to a minimum of (6), and in light of the experiment from Section 3.1, can be viewed as a direct way of constricting or shrinking the x-axis of Figure 2(right). Indeed, for sufficiently large $K$, once $\boldsymbol{w}$ has converged, $\boldsymbol{x}$ will immediately follow. But is this necessarily a desirable course of action?

Subplots (b)$-$(e) of Figure 3 show the support patterns of the $\hat{\boldsymbol{x}}$ estimate obtained via this procedure using $K \in \{1, 10, 100, 1000\}$. Only the $K = 100$ case produces a perfect recovery with matching support. It therefore follows that inner-loop iterations, when interpreted as a tunable sequence, have the potential to improve performance. Of course in advance we have no way of knowing what the best $K$ might be. But at least we do know that the $K = 1$ case which emerges from the original LSTM template need not be optimal.

## 4 Extension to Gated Feedback Networks

The discrepancy in convergence rates described in Section 3.1 occurs in part because the gate and cell updates do not fully solve the inner-loop weighted $\ell_1$ optimization needed to compute a globally optimal $\boldsymbol{x}^{(t+1)}$ given $\boldsymbol{w}^{(t)}$ (see Section 2.3). Varying the number of inner-loop iterations, meaning additional executions of (8)$-$(11) with $\boldsymbol{w}^{(t)}$ fixed, is one heuristic for normalizing across different trajectory frequencies, but this requires additional computational overhead, and prior knowledge is needed to micro-manage iteration counts for either efficiency or final estimation quality, the latter sensitivity being exposed in Section 3.2. For example, navigating around suboptimal local minima could require adaptively adjusting the number inner-loop iterations in subtle, non-obvious ways, with no discernible rule of thumb for enhancing solution quality. We therefore arrive at an unresolved state of affairs:

1. The latent variables which define SBL iterations can potentially follow optimization trajectories with radically different time scales, or both long- and short-term dependencies.

2. But there is no intrinsic mechanism within the SBL framework itself for *automatically* calibrating the differing time scales for optimal performance.[5]

These same issues are likely to arise in other non-convex multi-loop optimization algorithms as well. It therefore behooves us to consider a broader family of model structures that can adapt these scales in a data-dependent fashion.

In addressing this fundamental problem, we make the following key observation: *If the trajectories of various latent variables can be interpreted as activations passing through an RNN with both long- and short-term dependencies, then in developing a pipeline for optimizing such trajectories it makes sense to consider learning deep architectures explicitly designed to adaptively model such characteristic sequences.* Interestingly, in the context of sequence prediction, the *clockwork* RNN (CW-RNN) has been proposed to cope with temporal dependencies engaged across multiple scales [29]. As shown next however, the CW-RNN enforces dynamics synced to pre-determined clock rates exactly analogous to the fixed, manual schedule for terminating inner-loops (i.e., choosing $K$) in existing multi-loop iterative algorithms such as SBL. So we are back at our starting point. However, later in Section 4.2 we consider an alternative architecture to automate this process.

### 4.1 Clockwork Networks and Fixed Inner-Loop Iterations

In the context of sequence prediction, the *clockwork recurrent neural network* (CW-RNN) has been proposed to cope with temporal dependencies engaged across multiple scales [29]. In its most basic form, the CW-RNN begins with input, hidden, and output layers which, just like a regular RNN, are defined by

$$\boldsymbol{h}^{(t+1)} = f_H\left(\boldsymbol{W}_H \cdot \boldsymbol{h}^{(t)} + \boldsymbol{W}_I \cdot \boldsymbol{z}^{(t)}\right) \tag{15}$$

$$\boldsymbol{v}^{(t+1)} = f_O\left(\boldsymbol{W}_O \cdot \boldsymbol{h}^{(t+1)}\right), \tag{16}$$

where $\boldsymbol{z}^{(t)}$ is an input vector at time $t$, $\boldsymbol{h}^{(t)}$ represents hidden layer activations, $\boldsymbol{v}^{(t+1)}$ the output, and $\{\boldsymbol{W}_I, \boldsymbol{W}_H, \boldsymbol{W}_O\}$ are input, hidden, and output weight matrices respectively. Likewise, $f_H$ and $f_O$ are the corresponding nonlinear activation functions. What differentiates the CW-RNN from this vanilla structure, is that $\boldsymbol{W}_I$ and $\boldsymbol{W}_H$ are each partitioned into $g$ different temporarlly-varying block-rows[6] as

$$\boldsymbol{W}_I = \begin{bmatrix} \boldsymbol{W}_{I_1}^{(t)} \\ \vdots \\ \boldsymbol{W}_{I_g}^{(t)} \end{bmatrix}, \quad \boldsymbol{W}_H = \begin{bmatrix} \boldsymbol{W}_{H_1}^{(t)} \\ \vdots \\ \boldsymbol{W}_{H_g}^{(t)} \end{bmatrix}, \tag{17}$$

which naturally defines a corresponding segmentation of the hidden variables as

$$\boldsymbol{h}^{(t)} = \begin{bmatrix} \boldsymbol{h}_1^{(t)} \\ \vdots \\ \boldsymbol{h}_g^{(t)} \end{bmatrix} \tag{18}$$

such that, assuming separable nonlinearities,

$$\boldsymbol{h}_i^{(t+1)} = f_H\left(\boldsymbol{W}_{H_i}^{(t)} \cdot \boldsymbol{h}^{(t)} + \boldsymbol{W}_{I_i}^{(t)} \cdot \boldsymbol{z}^{(t)}\right), \; \forall i = 1, \ldots, g. \tag{19}$$

Additionally, each block is assigned and 'update period' $T_i$ that governs the structure across each time step $t$ via

$$\boldsymbol{W}_{I_i}^{(t)} = \begin{cases} \widetilde{\boldsymbol{W}}_{I_i} & \text{for } (t \bmod T_i) = 0 \\ [\mathbf{0}_1, \ldots, \mathbf{0}_g] & \text{otherwise} \end{cases} \tag{20}$$

and

$$\boldsymbol{W}_{H_i}^{(t)} = \begin{cases} \widetilde{\boldsymbol{W}}_{H_i} & \text{for } (t \bmod T_i) = 0 \\ [\mathbf{0}_1, \ldots, \mathbf{0}_{i-1}, \boldsymbol{I}, \mathbf{0}_{i+1}, \ldots, \mathbf{0}_g] & \text{otherwise.} \end{cases} \tag{21}$$

[5]Note that neither extreme, i.e., a single inner-loop iteration as with $K = 1$, or $K$ sufficiently large for full inner-loop convergence, need necessarily be optimal per the results from 3.2.

[6]A column-wise block structure may also be assumed if desired; however, this is not required for what follows herein.

In brief, these weight expressions ensure that for all $i$ we have

$$\boldsymbol{h}_i^{(t+1)} = \begin{cases} f_H\left(\widetilde{\boldsymbol{W}}_{H_i} \cdot \boldsymbol{h}^{(t)} + \widetilde{\boldsymbol{W}}_{I_i} \cdot \boldsymbol{z}^{(t)}\right) & \text{for } (t \bmod T_i) = 0 \\ \boldsymbol{h}_i^{(t)} & \text{otherwise.} \end{cases} \tag{22}$$

This formulation allows the CW-RNN to handle different temporal features by assigned different $T_i$ to different blocks. For example, a block designed to model high-frequency dynamics may assume $T_i = 1$, while slowly-varying components can be captured using $T_i \gg 1$. The latter implies that for most iterations, the block hidden state $\boldsymbol{h}_i^{(t)}$ is not updated allowing for hard-coded long-term memory of such low-frequency dynamics.

This prescription exactly reflects the basic anatomy of an algorithm with $g$ nested loops, each loop being characterized by its own set of latent variables $\boldsymbol{h}_i^{(t)}$. As a simple example, consider the SBL updates equipped with an inner-loop as in Section 3.2. If we define $\boldsymbol{h}_1^{(t)} = \boldsymbol{x}^{(t)}$, $\boldsymbol{h}_2^{(t)} = \boldsymbol{w}^{(t)}$, $\boldsymbol{z}^{(t)} = \boldsymbol{y}$, adopt $T_1 = 1$ and $T_2 = K$, and relabel the iteration numbers via a single consistent index (i.e., we collapse $k$ and $t$ into a single index), then $\boldsymbol{w}^{(t)}$ will only be updated once every $T_2$ time-steps, while $\boldsymbol{x}^{(t)}$ will be updated at all $t$, and the basic scheduling is identical. The only difference is that the layer-wise filters and nonlinearities are somewhat more specialized for the SBL context.

### 4.2 Automatically Scheduling Inner-Loops via Gated Feedback

The gated feedback RNN (GF-RNN) [15] was recently developed to update the CW-RNN with an additional set of gated connections that, in effect, allow the network to learn its own clock rates. In brief, the GF-RNN involves stacked LSTM layers (or somewhat simpler gated recurrent unit (GRU) layers [14]), that are permitted to communicate bilaterally via additional, data-dependent gates that can open and close on different time-scales. In the context of SBL, this means that we no longer need strain a specialized LSTM structure with the burden of coordinating trajectory dynamics. Instead, we can stack layers that are, at least from a conceptual standpoint, designed to reflect the different dynamics of disparate variable sets such as $\boldsymbol{w}^{(t)}$ or $\boldsymbol{x}^{(t)}$.

In doing so, we are then positioned to learn new SBL update rules from training pairs $\{\boldsymbol{y}, \boldsymbol{x}^*\}$ as described previously. At the very least, this structure should include SBL-like iterations within its capacity, but of course it is also free to explore something even better. We describe our detailed implementation next.

### 4.3 Network Design and Training Protocol

We stack two gated recurrent layers loosely designed to mimic the relatively fast SBL adaptation to basic correlation structure, and slower resolution of final support patterns and coefficient estimates. These layers are formed from a LSTM base architecture (or sometimes a GRU for comparison purposes). For the final output layer we adopt a multi-label classification loss for predicting supp$[\boldsymbol{x}^*]$, which is the well-known 'NP-hard' part of sparse estimation (determining final coefficient amplitudes just requires least squares). Full network details are deferred to Appendix A, including special modifications to handle complex data as required by DOA applications.

For a given dictionary $\boldsymbol{\Phi}$ a separate network must be trained via SGD, to which we add a unique extra dimension of randomness via an online stochastic data-generation strategy. In particular, to create samples in each mini-batch, we first generate a vector $\boldsymbol{x}^*$ with random support pattern and nonzero amplitudes. We then compute $\boldsymbol{y} = \boldsymbol{\Phi}\boldsymbol{x}^* + \boldsymbol{\epsilon}$, where $\boldsymbol{\epsilon}$ is a small Gaussian noise component. This $\boldsymbol{y}$ forms a training input sample, while supp$[\boldsymbol{x}^*]$ represents the corresponding labels. For all mini-batches, novel samples are drawn, which we have found boosts performance considerably over the fixed training sets used by current DNN approaches to sparse estimation (see Appendix A).

## 5 Experiments

This section presents experiments involving synthetic data and two applications.

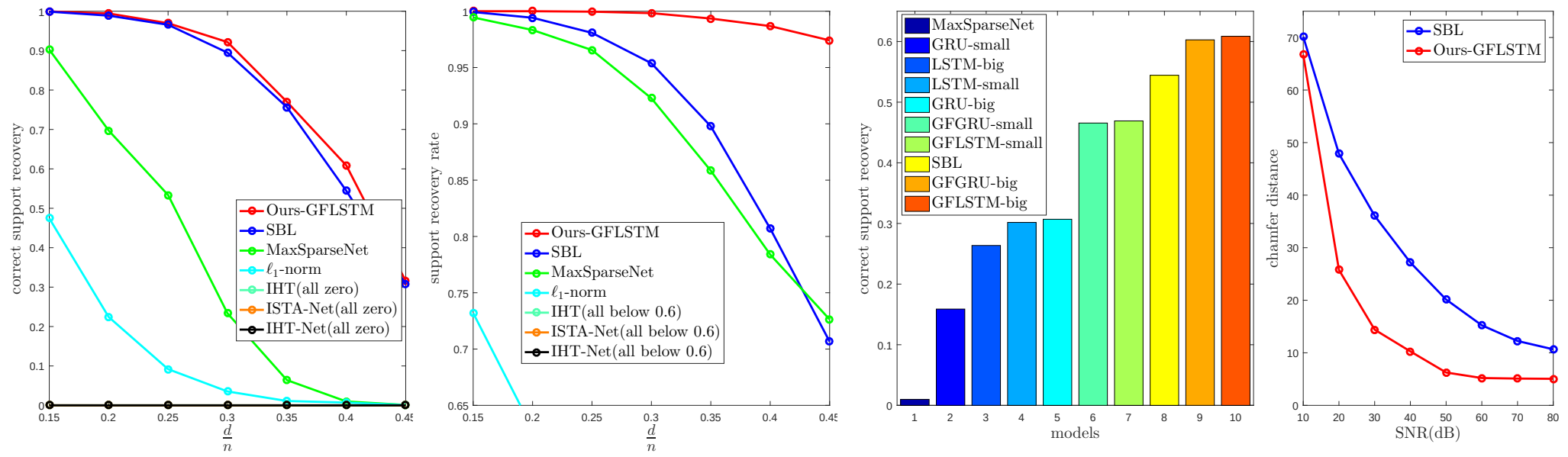

(a) *Strict Accuracy* (b) *Loose Accuracy* (c) *Architecture Comparisons* (d) *DOA*

Figure 4: Plots (a), (b), and (c) show sparse recovery results involving synthetic correlated dictionaries. Plot (d) shows Chamfer distance-based errors [8] from the direction-of-arrival (DOA) experiment.

### 5.1 Evaluations via Synthetic Correlated Dictionaries

To reproduce experiments from [2], we generate correlated synthetic features via $\mathbf{\Phi} = \sum_{i=1}^{n} \frac{1}{i^2} \boldsymbol{u}_i \boldsymbol{v}_i^\top$, where $\boldsymbol{u}_i \in \mathbb{R}^n$ and $\boldsymbol{v}_i \in \mathbb{R}^m$ are drawn iid from a unit Gaussian distribution, and each column of $\mathbf{\Phi}$ is subsequently rescaled to unit $\ell_2$ norm. Ground truth samples $\boldsymbol{x}^*$ have $d$ nonzero elements drawn randomly from $\mathcal{U}[-0.5, 0.5]$ excluding the interval $[-0.1, 0.1]$. We use $n$=20, $m$=100, and vary $d$, with larger values producing a much harder combinatorial estimation problem (exhaustive search is not feasible here). All algorithms are presented with $\boldsymbol{y}$ and attempt to estimate supp$[\boldsymbol{x}^*]$. We evaluate using *strict accuracy*, meaning percentage of trials with exact support recovery, and *loose accuracy*, which quantifies the percentage of true positives among the top $n$ 'guesses' (i.e., largest predicted outputs).

Figures 4(a) and 4(b) evaluate our model, averaged across $10^5$ trials, against an array of optimization-based approaches: SBL [38], $\ell_1$ norm minimization [5], and IHT [6]; and existing learning-based DNN models: an ISTA-inspired network [23], an IHT-inspired network [39], and the best maximal sparsity net (MaxSparseNet) from [2] (detailed settings in Appendix A). With regard to strict accuracy, only SBL is somewhat competitive with our approach and other learning-based models are much worse; however, using loose accuracy our method is far superior than all others. Note that this is the first approach we are aware of in the literature that can convincingly outperform SBL recovering sparse solutions when a heavily correlated dictionary is present, and we hypothesize that this is largely possible because our design principles were directly inspired by SBL itself.

To isolate architectural factors affecting performance we conducted ablation studies: (i) with or without gated feedback, (iii) LSTM or GRU cells, and (iii) small or large ($4\times$) model size; for each model type, the small and respectively large versions have roughly the same number of parameters. Appendix D also contains a much broader set of self-comparison studies. Figure 4(c), which shows strict accuracy results with $d/n = 0.4$, indicates the importance of gated feedback and to a lesser degree network size, while LSTM and GRU cells perform similarly as expected.

### 5.2 Practical Application I: Direction-of-Arrival (DOA) Estimation

DOA estimation is a fundamental problem in sonar/radar processing [32]. Given an array of $n$ omnidirectional sensors with $d$ signal waves impinging upon them, the objective is to estimate the angular direction of the wave sources with respect to the sensors. For certain array geometries and known propagation mediums, estimation of these angles can be mapped directly to solving (2) in the complex domain. In this scenario, each column of $\mathbf{\Phi}$ represents the sensor array output (a point in $\mathbb{C}^n$) from a hypothetical source with unit strength at angular location $\theta_i$, and can be computed using wave progagation formula [32]. The entire dictionary can be constructed by concatenating columns associated with angles forming some spacing of interest, e.g., every $1^\circ$ across a half circle, and will be highly correlated. Given measurements $\boldsymbol{y} \in \mathbb{C}^n$, we can solve (2), with $\lambda$ reflecting the noise level. The indexes of nonzero elements of $\boldsymbol{x}^*$ will then reveal the angular locations/directions of putative sources.

Recently SBL-based algorithms have produced state-of-the-art results solving the DOA problem [17, 22, 45], and we compare our approach against SBL here. We apply a typical experimental design from the literature involving a uniform linear array with $n = 10$ sensors; see Appendix

B for background and details on how to compute $\boldsymbol{\Phi}$ and specifics on how to adapt and train our GFLSTM using complex data. Four sources are then placed in random angular locations, with nonzero coefficients at $\{\pm 1 \pm i\}$, and we compute measurements $\boldsymbol{y} = \boldsymbol{\Phi}\boldsymbol{x}^* + \boldsymbol{\epsilon}$, with $\boldsymbol{\epsilon}$ chosen from a complex Gaussian distribution to produce different SNR. Because the nonzero positions in $\boldsymbol{x}^*$ now have physical meaning, we apply the Chamfer distance [8] as the error metric, which quantifies how close we are to true source locations (lower is better). Figure 4(d) displays the results, where our learned network outperforms SBL across a range of SNR values.

### 5.3 Practical Application II: 3D Geometry Recovery via Photometric Stereo

Photometric stereo represents another application domain whereby approximately solving (2) using SBL has recently produced state-of-the-art results [27]. The objective here is to recover the 3D surface normals of a given scene using $r$ images taken from a single camera using different lighting conditions. Under the assumption that these images can be approximately decomposed into a diffuse Lambertian component and sparse corruptions such as shadows and specular highlights, then surface normals at each pixel can be recovered using (2) to isolate these sparse factors followed by a final least squares post-processing step [27]. In this context, $\boldsymbol{\Phi}$ is constructed using the known camera and lighting geometry, and $\boldsymbol{y}$ represents intensity measurements for a given pixel across images projected onto the nullspace of a special transposed lighting matrix (see Appendix C for more details and our full experimental design). However, because a sparse regression problem must be computed for every pixel to recovery the full scene geometry, a fast, efficient solver is paramount.

Table 1: Photometric stereo results

| Algorithm | Average angular error (degrees) | | | | | | Runtime (sec.) | | | | | |
|---|---|---|---|---|---|---|---|---|---|---|---|---|
| | Bunny | | | Caesar | | | Bunny | | | Caesar | | |
| | r=10 | r=20 | r=40 | r=10 | r=20 | r=40 | r=10 | r=20 | r=40 | r=10 | r=20 | r=40 |
| SBL | 4.02 | 1.86 | **0.50** | 4.79 | 2.07 | **0.34** | 35.46 | 22.66 | 32.20 | 86.96 | 64.67 | 90.48 |
| MaxSparseNet | 1.48 | 1.95 | 1.20 | 3.51 | 2.51 | 1.18 | 0.90 | 0.87 | 0.92 | 2.13 | 2.12 | 2.20 |
| Ours | **1.35** | **1.55** | 1.12 | **2.39** | **1.80** | 0.60 | **0.63** | **0.67** | **0.85** | **1.48** | **1.70** | **2.08** |

We compare our GFLSTM model against both SBL and the MaxSparseNet [2] (both of which outperform other existing methods). Tests are performed using the 32-bit HDR gray-scale images of objects 'Bunny' ($256 \times 256$) and 'Caesar' ($300 \times 400$) as in [27]. For (very) weakly-supervised training data, we apply the same approach as before, only we use nonzero magnitudes drawn from a Gaussian, with mean and variance loosely tuned to the photometric stereo data, consistent with [2]. Results are shown in Table 1, where we observe in all cases the DNN models are faster by a wide margin, and in the hard cases cases (smaller $r$) our approach produces the lowest angular error. The only exception is with $r = 40$; however, this is a quite easy scenario with so many images such that SBL can readily find a near optimal solution, albeit at a high computational cost. See Appendix C for error surface visualizations.

## 6 Conclusion

We have demonstrated that gated recurrent nets carefully patterned to reflect the multi-scale optimization trajectories of multi-loop SBL iterations can lead to a considerable boost in both accuracy and efficiency. Note that simpler first-order, gradient descent-style algorithms can be ineffective when applied to sparsity-promoting energy functions with a combinatorial number of bad local optima and highly concave or non-differentiable surfaces in the neighborhood of minima. Moreover, implementing smoother approximations such as SBL with gradient descent is impractical since each gradient calculation would be prohibitively expensive. Therefore, recent learning-to-learn approaches such as [1] that rely on gradient descent are difficult to apply in the present setting.

## 7 Appendix A: Modeling and Training Details

We first describe the basic gated feedback RNN structure, following by our particular model architecture including extensions to handle complex data. We conclude with training details and experimental settings.

### 7.1 Gated Feedback RNN Structure

The gated feedback RNN cell [15] is a key component of our model. Detailed computing flows for a gated feedback LSTM cell (GFLSTM), which represents one particular specialization that is used in all our experiments, follow as

$$\begin{aligned}
\boldsymbol{c}_j^{(t)} &= \boldsymbol{f}_j^{(t)} \odot \boldsymbol{c}_j^{(t-1)} + \boldsymbol{i}_j^{(t)} \odot \tilde{\boldsymbol{c}}_j^{(t)} \\
\boldsymbol{h}_j^{(t)} &= \boldsymbol{o}_j^{(t)} \odot \text{Tanh}(\boldsymbol{c}_j^{(t)}) \\
\boldsymbol{i}_j^{(t)} &= \sigma(\boldsymbol{W}_{ij}\boldsymbol{a}_j^{(t)} + \boldsymbol{U}_{ij}\boldsymbol{h}_j^{(t-1)}) \\
\boldsymbol{f}_j^{(t)} &= \sigma(\boldsymbol{W}_{fj}\boldsymbol{a}_j^{(t)} + \boldsymbol{U}_{fj}\boldsymbol{h}_j^{(t-1)}) \\
\boldsymbol{o}_j^{(t)} &= \sigma(\boldsymbol{W}_{oj}\boldsymbol{a}_j^{(t)} + \boldsymbol{U}_{oj}\boldsymbol{h}_j^{(t-1)}) \\
\boldsymbol{g}_{i\to j}^{(t)} &= \sigma(\boldsymbol{W}_{gj}\boldsymbol{a}_j^{(t)} + \boldsymbol{U}_{g_{i\to j}}\boldsymbol{H}^{(t-1)}) \\
\tilde{\boldsymbol{c}}_j^{(t)} &= \text{Tanh}(\boldsymbol{W}_{cj-1\to j}\boldsymbol{h}_{j-1}^{(t)} + \sum_{i=1}^{r} \boldsymbol{g}_{i\to j}^{(t)} \odot \boldsymbol{U}_{ci\to j}\boldsymbol{h}_i^{(t-1)}),
\end{aligned} \tag{23}$$

where $r$ is the number of stacked LSTM cells, subscript $j$ is the index for LSTM cell in the stack, while superscript $(t)$ indicates the time point. Therefore $\boldsymbol{h}_j^{(t)}$ and $\boldsymbol{c}_j^{(t)}$ denote the *hidden state* and *memory cell* of the *j-th* LSTM unit in the stack at time $t$. And we denote $\boldsymbol{a}_j^{(t)}$ as the input of the *j-th* LSTM cell, such that $\boldsymbol{a}_j^{(t)} = \boldsymbol{h}_{j-1}^{(t)} (\forall j > 1)$ and $\boldsymbol{a}_1^{(t)} = \boldsymbol{y}$. Besides conventional designs like an *input gate* $\boldsymbol{i}_j^{(t)}$, *forget gate* $\boldsymbol{f}_j^{(t)}$, and *output gate*, $\boldsymbol{o}_j^{(t)}$, the stack of GFLSTM cells also includes an extra *global gate* computed from input $\boldsymbol{a}_j^{(t)}$ and $\boldsymbol{H}^{(t-1)} = [\boldsymbol{h}_1^{(t-1)}, .., \boldsymbol{h}_r^{(t-1)}]$, the concatenation of all the hidden states from the previous time step $t-1$. Each $\boldsymbol{g}_{i\to j}^{(t)}$ controls the flow from $\boldsymbol{h}_i^{(t-1)}$ to $\boldsymbol{h}_j^{(t)}$, that is, the layer cross feedback. To make it concise, we can denote the whole computing flow of these $r$ LSTM cells using the function $f_{GFLSTM}$ as

$$\begin{aligned}
&f_{GFLSTM}(\boldsymbol{H}^{(t-1)}, \boldsymbol{y}; \boldsymbol{\theta}_{GFLSTM}) = [\boldsymbol{q}^{(t)}, \boldsymbol{H}^{(t)}] \\
&\boldsymbol{\theta}_{GFLSTM} = [\boldsymbol{W}_{ij}, \boldsymbol{W}_{fj}, \boldsymbol{W}_{oj}, \boldsymbol{U}_{ij}, \boldsymbol{U}_{fj}, \boldsymbol{U}_{oj}, \boldsymbol{W}_{gj}, \boldsymbol{U}_{g_{i\to j}}, \boldsymbol{W}_{cj-1\to j}, \boldsymbol{U}_{ci\to j}] \\
&\boldsymbol{q}^{(t)} = \boldsymbol{h}_r^{(t)}.
\end{aligned} \tag{24}$$

### 7.2 Proposed Model Architecture and Extensions

**Basic Model:** Although our model consists of RNN cells, once we fix the number of unfolding steps, it essentially becomes a feed-forward network. As shown in Figure 5, during the forward stage, the input is broadcasted to the lowest RNN cell at each unrolled step. After the model generates its outputs at each unrolled step, all of these outputs will be concatenated and fed into a *fully connected* layer to produce the final prediction. Since we opt to predict the $supp[\boldsymbol{x}^*] = \{i : x_i^* \neq 0\}$, we view the problem as an multi-label classification task and append a *softmax* layer on top of the fully connected layer. We formalize this process as

$$\begin{aligned}
[\boldsymbol{q}^{(t)}, \boldsymbol{H}^{(t)}] &= f_{rnn}(\boldsymbol{H}^{(t-1)}, \boldsymbol{y}; \boldsymbol{\theta}_{rnn}) \\
\boldsymbol{p} &= f_{pred}([\boldsymbol{q}^{(1)}, \boldsymbol{q}^{(2)}, .., \boldsymbol{q}^{(T)}]; \boldsymbol{\theta}_{pred}) \\
f_{pred}(\boldsymbol{q}^{(all)}, \boldsymbol{\theta}_{pred}) &= softmax(\boldsymbol{W}_{pred}\boldsymbol{q}^{(all)} + \boldsymbol{b}_{pred}) \\
\boldsymbol{q}^{(all)} &= [\boldsymbol{q}^{(1)}, \boldsymbol{q}^{(2)}, .., \boldsymbol{q}^{(T)}],
\end{aligned} \tag{25}$$

where $\boldsymbol{\theta}_{rnn}, \boldsymbol{\theta}_{pred} = [\boldsymbol{W}_{pred}, \boldsymbol{b}_{pred}]$ are the parameters of the RNN units and the fully connected layer respectively. $\boldsymbol{q}^{(t)}$, $\boldsymbol{h}^{(t)}$ denote the output of the RNN units and hidden state at each time step $t$. In practice, we simply take the RNN's top layer hidden states as its output $\boldsymbol{q}^{(t)}$. $f_{rnn}$ represents the forward process of the RNN, which is defined by the exact structure of the RNN-cell.

**Complex Value Extension:** In many real applications of sparse recovery, the format of the inputs may vary. For example, the inputs to the DOA problem of interest are complex numbers. We propose to deal with complex-value inputs by what we call *model complexification*. Specifically, RNN units consist of matrix multiplication and non-linear activations, both of which have their complex value counterparts. Thus we propose to use complex-value operations in the RNN units before finally concatenating the real and imaginary part of the outputs as the feature for the final prediction. This method is inspired by SBL, which handles real and complex value inputs with the same operator. We argue that this method is better than simply concatenating the real and imaginary parts of the input and using a regular real-valued RNN (of double the size) for prediction, since in this way the links between real and imaginary parts of a complex number are broken and therefore the RNN may potentially have to learn these links by itself, which leads to unnecessary learning difficulties.

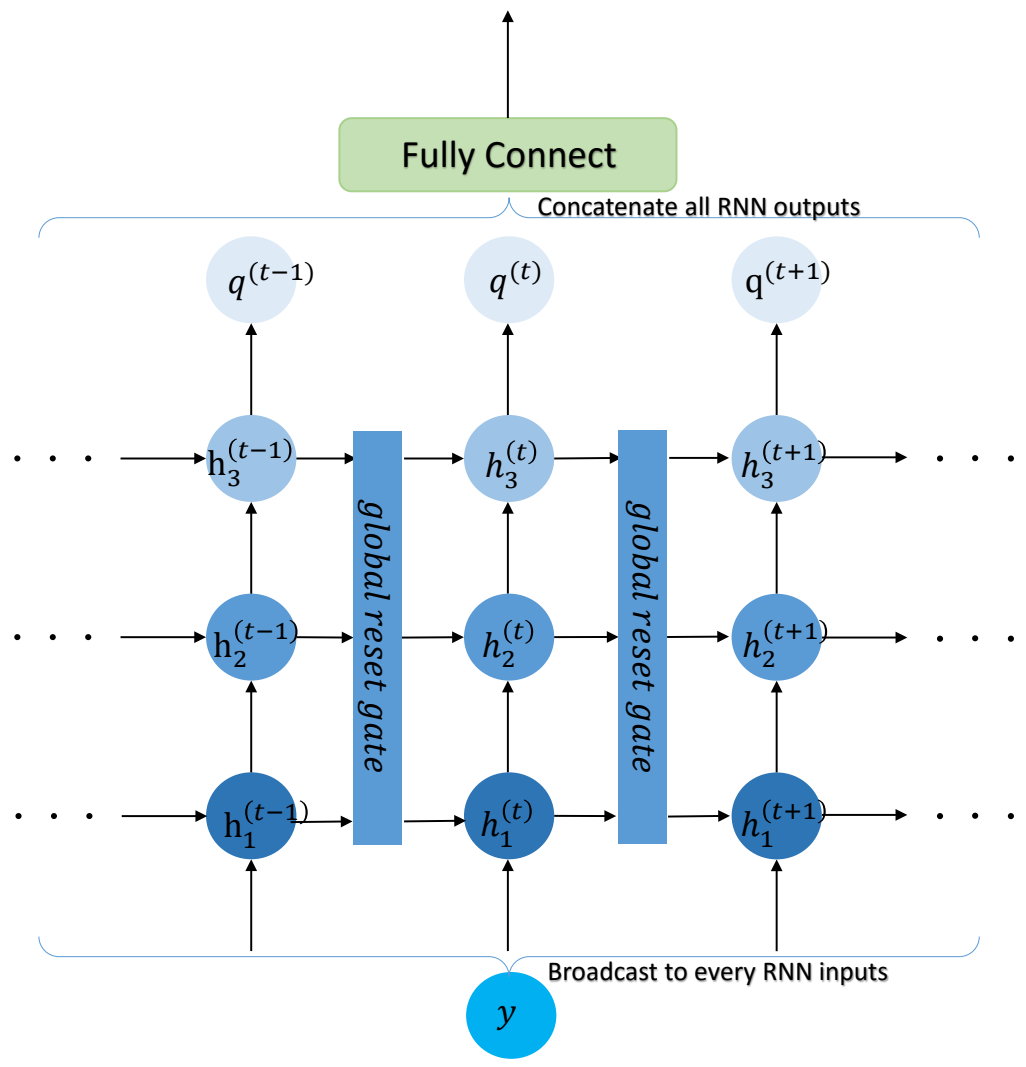

Figure 5: Proposed model architecture using gated feedback LSTM cell

### 7.3 Training Details

We apply a unified training framework for all different approaches. In our experiments, models are implemented using Torch7 and experiments are run on a single NVIDIA Tesla K40M GPU card.

**Training Hyperparameters:** To provide consistency with the concept of epoch from [2], our models are trained by $600000/250 = 2400$ batches with *batch size* equal to 250. Typically, with 400 epochs (or 800 epochs in some extreme cases) of RMSprop optimization, we converge to a satisfactory performance level, with a default initial learning rate of 0.002, factored by 0.25 every 50 epochs after the first 250 epochs of training.

**Model Hyperparameters:** As for model architecture, there exists the following hyperparameters: number of RNN hidden units $h$, number of stacked RNN layers $r$ and number of RNN-cell unfold steps $T$. In almost all of our experiments, we control model capacity mainly by the size of hidden states with fixed layer number $r = 2$ and unfold steps $T = 11$. In section 10.2, we provide more detailed ablation studies on how the number of RNN layers, unroll steps and hidden units affects the performance.

**A Useful Training Heuristic:** When training with a fixed-sized dataset, as existing learning approaches to sparse estimation do [23, 39, 2], there is always the risk of overfitting. The gap between the error on training and validation sets with a fixed dataset is shown via the blue curves in Figure 6(a)) on a representative learning problem. However, since we are free to generate as much training data as we want in the sparse estimation context, at every epoch we can always use a new, unseen batch. This simple strategy completely closes the gap (the red curves) with negligible computational overhead. Figure 6(b) displays the resulting improvement on performance, as measured by the percentage of trials whereby the entire support pattern is correctly estimated (i.e. *strict accuracy*).

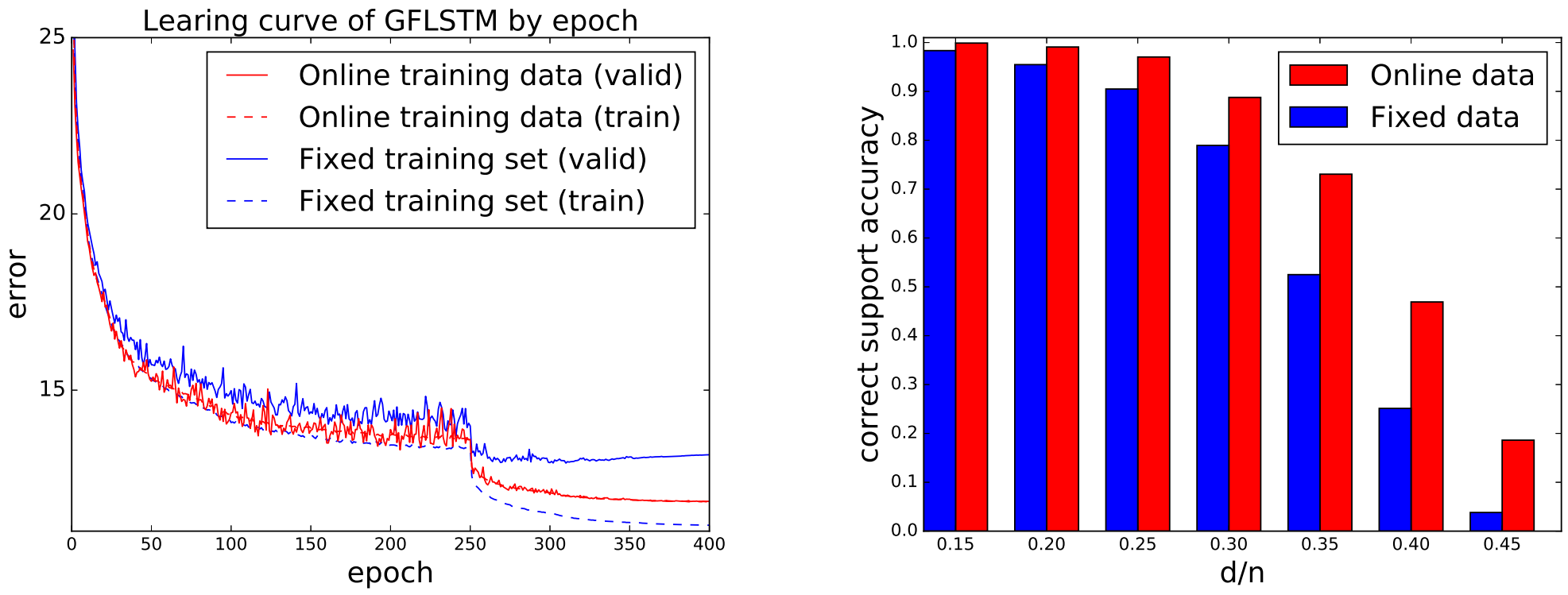

(a) (b)

Figure 6: Demonstration of online training heuristic benefits

# 8 Appendix B: Experimental Details for Direction-of-Arrival (DOA) Estimation

This appendix contains background information, followed by experimental and training details.

## 8.1 Background

Direction-of-arrival (DOA) estimation for sonar and radar application can be formulated by the observation model

$$\boldsymbol{y}(t) = \sum_{k=1}^{d} s_k(t) f(\theta_k^*) + \boldsymbol{\epsilon}(t), \tag{26}$$

where $\boldsymbol{y}(t) \in \mathbb{C}^n$ is the measured sonor/radar signal at time $t$, $f : \mathbb{R} \rightarrow \mathbb{C}^n$, and $d$ is number of source waveforms whose magnitudes are $\boldsymbol{s}(t) = [s_1(t), ..., s_d(t)]^\top \in \mathbb{C}^d$ and angular locations are $\boldsymbol{\theta}^* = [\theta_1^*, ..\theta_d^*]^\top$ [32]. Although the location space $\Theta$ might be continuous, we may approximate it by a fixed sampling grid $\boldsymbol{\theta} = [\theta_1, .., \theta_m]$. Then the problem can be rewritten as the alternative observation model

$$\boldsymbol{y}(t) = \sum_{i=1}^{m} x_i(t)\boldsymbol{\phi}_i + \boldsymbol{\epsilon}(t) = \boldsymbol{\Phi}\boldsymbol{x}(t) + \boldsymbol{\epsilon}(t), \tag{27}$$

where $\boldsymbol{\Phi} = [\boldsymbol{\phi}_1, .., \boldsymbol{\phi}_m]$, $\boldsymbol{\phi}_i \triangleq f(\theta_i)$, and $\boldsymbol{x}(t) = [x_1(t), ..., x_m(t)]^\top$. With sufficient resolution provided, we assume every $\theta_k^*$ is contained in $\boldsymbol{\theta}$ such that $\boldsymbol{s}(t)$ becomes a collection of $d$ non-zero entries in $\boldsymbol{x}(t)$. Finally, the DOA estimation problem boils down to solving $\min_{\boldsymbol{x}} \|\boldsymbol{y}-\boldsymbol{\Phi}\boldsymbol{x}\|_2^2+\lambda\|\boldsymbol{x}\|_0$, whereby nonzero elements in $\boldsymbol{x}^*$ will (approximately) correspond with locations $i$ whereby $\boldsymbol{\theta}_i \approx f(\theta_k^*)$ for some $k$.

## 8.2 Experimental Design

We make some natural assumptions in our experiment. First we consider the narrowband, far-field case which implies that incoming waves are approximately planar and each source emanates from a single point. Furthermore, we assume our sensors are arranged having a linear, uniformly spaced array geometry, i.e., *uniform linear array*(ULA), and a known propagation medium. Then the measurement vector $\boldsymbol{y}(t)$ obtained by sensors at time $t$ is given by (26), where the non-linear function $f$ is

$$\begin{aligned} f(\theta) &= \left[e^{i\omega_0\Delta_1(\theta)}, .., e^{i\omega_0\Delta_n(\theta)}\right]^\top \quad \text{with} \\ \omega_0\Delta_j(\theta) &= 2\pi(j-1)\frac{Dcos(\theta)}{\lambda_0}, \forall j = 1, .., n, \end{aligned} \tag{28}$$

and $\omega_0$, $\lambda_0$ are the central temporal frequency and the wavelength of signals respectively. Also, $\Delta_j(\theta)$ is the array-geometry-dependent time delay between the first sensor and the $j$-th sensor for a given angle $\theta \in [0, \pi]$, while $D$ is the distance between two nearby sensors in the ULA.

**Settings:** In our experiments, we set $m = 180$ allowing an angular resolution of $1°$ over the half circle, and $n = 10$ sensors with $D = 0.5\lambda_0$. The dictionary $\mathbf{\Phi}$ is constructed via (26) and (28) such that the $i$-th column represents the sensor array output from a hypothetical source of unit strength at angular location $\theta_i$. The number of different sources $d$ is set to 4, which is represents a quite challenging problem with only 10 sensors; most sparse estimation algorithms will fail in this regime. Then we randomly pick four different source directions with magnitudes $\{\pm 1 \pm i\}$. Finally, a measurement vector $\boldsymbol{y} = \mathbf{\Phi}\boldsymbol{x} + \boldsymbol{\epsilon}$ is calculated with complex Gaussian noise added to maintain a given signal noise ratio (SNR).

**Metric:** We apply the symmetric Chamfer distance[8] to evaluate the estimation quality with respect to the ground truth source directions. This distance is given by $\boldsymbol{\theta}^* = \{\theta_1^*, ..\theta_d^*\}$ and the prediction $\hat{\boldsymbol{\theta}} = \{\hat{\theta}_1, ..\hat{\theta}_d\}$

$$dist(\boldsymbol{\theta}^*, \hat{\boldsymbol{\theta}}) = \sum_{\theta_1 \in \boldsymbol{\theta}^*} \min_{\theta_2 \in \hat{\boldsymbol{\theta}}} |\theta_1 - \theta_2| + \sum_{\theta_2 \in \hat{\boldsymbol{\theta}}} \min_{\theta_1 \in \boldsymbol{\theta}^*} |\theta_1 - \theta_2|. \tag{29}$$

**Training Details:** For DOA experiments, our model has LSTM cells with 200 hidden units and is trained 400 epoches following our default settings. For training data generation, we tried using noise levels in the intervals $[15dB, 30dB]$, $[20dB, 40dB]$, $[30dB, 60dB]$, and $[60dB, 80dB]$, and then chose the best results.

# 9 Appendix C: Experimental Details for 3D Geometry Recovery via Photometric Stereo

This appendix describes our photometric stereo experiments in more detail, including error surface visualizations.

## 9.1 Background

Photometric stereo represents a useful method for recovering high-resolution surface normals from a 3D scene using 2D images taken under $r$ different lighting conditions. One proposed model for the observation process at a single pixel is

$$\boldsymbol{o} = \rho \boldsymbol{L} \boldsymbol{n} + \boldsymbol{e}, \tag{30}$$

where the $r$ measurments are denoted $\boldsymbol{o} \in \mathbb{R}^r$, $\boldsymbol{n} \in \mathbb{R}^3$ denotes the true 3D surface normal, rows of $\boldsymbol{L} \in \mathbb{R}^{r \times 3}$ define lighting directions, $\rho$ is the diffuse albedo, acting here as a scalar multiplier, and $\boldsymbol{e}$ represents an aggregations of shadows, specular highlights, or other corrupting influences [28, 44]. If $\boldsymbol{e}$ were not present, then the surface normals can be uniquely determined using a simple least-squares fit. However, a more robust alternative involves solving

$$\min_{\tilde{\boldsymbol{n}}, \boldsymbol{e}} \|\boldsymbol{e}\|_0 \quad \text{s.t.} \quad \boldsymbol{o} = \boldsymbol{L}\tilde{\boldsymbol{n}} + \boldsymbol{e}, \tag{31}$$

where $\tilde{\boldsymbol{n}}$ is the surface normal rescaled by $\rho$, which is equivalent to computing [28]

$$\min_{\boldsymbol{e}} \|\boldsymbol{e}\|_0 \quad \text{s.t. } \text{Proj}_{\text{null}[L^\top]}(\boldsymbol{o}) = \text{Proj}_{\text{null}[L^\top]}(\boldsymbol{e}). \tag{32}$$

It can be shown that this formulation has the exact same structure as (2) in the limit $\lambda \to 0$, if we assume that $\boldsymbol{y} \triangleq \text{Proj}_{\text{null}[L^\top]}(\boldsymbol{o})$ and $\mathbf{\Phi}$ is defined such that $\mathbf{\Phi}\boldsymbol{e} = \text{Proj}_{\text{null}[L^\top]}(\boldsymbol{e})$.

## 9.2 Experimental Design

We test algorithms separately on two objects, ‘Bunny’ and ‘Caesar’ from [27]. First lighting conditions are generated whose directions are randomly selected from a hemisphere with the object placed at the center. Then 32-bit HDR gray-scale images of the object are rendered with foreground masks and a randomly chosen $\rho$, 0.64 for Bunny and 0.8 for Caesar. The resulting image resolution for Bunny is $(256 \times 256)$ while for Caesar it is $(300 \times 400)$. Given $\boldsymbol{L}$, we apply singular value decomposition to get $\boldsymbol{\Phi} = \text{Proj}_{\text{null}[L^\top]}$ and the ground truth error vector $\boldsymbol{e}^* = \boldsymbol{o} - \rho \boldsymbol{L}\boldsymbol{n}$.

Table 2: Attributes of our models used in producing Figure 4(c) results.

| model | Hidden Unit Size | #Parameters | Training Time(sec./epoch) | S-Acc |
|---|---|---|---|---|
| GRU-small | 320 | 1296740 | 98.314 | 0.1588 |
| LSTM-small | 272 | 1213220 | 119.605 | 0.3017 |
| GFGRU-small | 220 | 1285340 | 170.282 | 0.4651 |
| GFLSTM-small | 200 | 1209300 | 172.013 | 0.4691 |
| GRU-big | 680 | 4958660 | 234.024 | 0.3069 |
| LSTM-big | 600 | 5037700 | 312.642 | 0.2637 |
| GFGRU-big | 455 | 4903635 | 318.690 | 0.6028 |
| GFLSTM-big | 425 | 4864650 | 310.447 | 0.6087 |

For training, we have to synthesize candidate sparse errors $\boldsymbol{e}$ since there is no photometric stereo database for this purpose. We adopt the basic pipeline from [2] to accomplish this. First we draw a support pattern for $\boldsymbol{e}$ uniformly at random with cardinality $d$ sampled from the range $[d_1, d_2]$. Nonzero values of $\boldsymbol{e}$ are assigned iid random values from $\mathcal{N}(\mu_e, \sigma_e)$. Finally, we can naturally compute observations $\boldsymbol{y} = \boldsymbol{\Phi}\boldsymbol{e}$ which serve as network inputs. Although $d_1$,$d_1$,$\mu_e$, and $\sigma_e$ are all tunable, beyond this, no attempt is made to match the true outlier distributions encountered in applications of photometric stereo. After training on synthetic data, we directly apply the resulting model to the gray-scale images without any additional application-specific tuning. During the testing stage, for each surface point, we use our model to approximately solve (32). Since the network outputs a probability map for the outlier support set, we choose $k$ indices with the least probability as inliers and use them to compute $\boldsymbol{n}$ via least squares.

**Hyperparameters:** We conduct experiments under three situations using $r = 10, 20, 40$ images corresponding to $r$ different lighting conditions. As for model capacity, we set the size of hidden states of LSTM cells equal to $2r$. Other training settings remain default as in Section 7.3.

**Visual Results:** See Figure 7.

# 10 Appendix D: Additional Experiments and Self-Comparisons

We provide more evaluation details for generic sparse recovery problems, followed by a number of ablation studies.

## 10.1 Additional Details for Sparse Vector Recovery Evaluation

Table 2 lists all the important attributes of our self-comparison models from Figure 4(c) in the main paper. In terms of evaluation on generic problems, we define *strict accuracy*(s-acc) and *loose accuracy*(l-acc) via

$$\mathcal{S}_{gt} = \{j : x_j^* \neq 0\}, \quad \mathcal{S}_{pred}(d) = \{j : p_j \text{ is one of the } d \text{ largest outputs}\} \tag{33}$$

$$\text{s-acc} = \tfrac{1}{N} \textstyle\sum_{i=1}^N \mathbb{I}\left[\mathcal{S}_{gt}^i = \mathcal{S}_{pred}^i(d)\right], \quad \text{l-acc} = \tfrac{1}{N} \textstyle\sum_{i=1}^N \frac{\left|\mathcal{S}_{gt}^i \cap \mathcal{S}_{pred}^i(n)\right|}{d}, \tag{34}$$

where $N$ is the number of samples.

## 10.2 Ablation Study for Generic Sparse Estimation Problem

In Table 3, we list an ablation results of GFLSTM models with different hyperparameters for the $\frac{d}{n} = 0.4$ case. Enlarging capacity generally benefits the performance especially when the capacity is relatively small. However, the effectiveness and efficiency of changing hidden size, LSTM layers, or number of unrolling steps varies. Stacking too many LSTM layers is the least efficient way the enlarge the model capacity considering the trade-off between training time and performance improvement. As for unrolling steps, insufficient steps (say under 10) by no means impairs the models ability while excessive unrolling is a waste of computation. And hidden size is a quite effective way to control the model capacity.

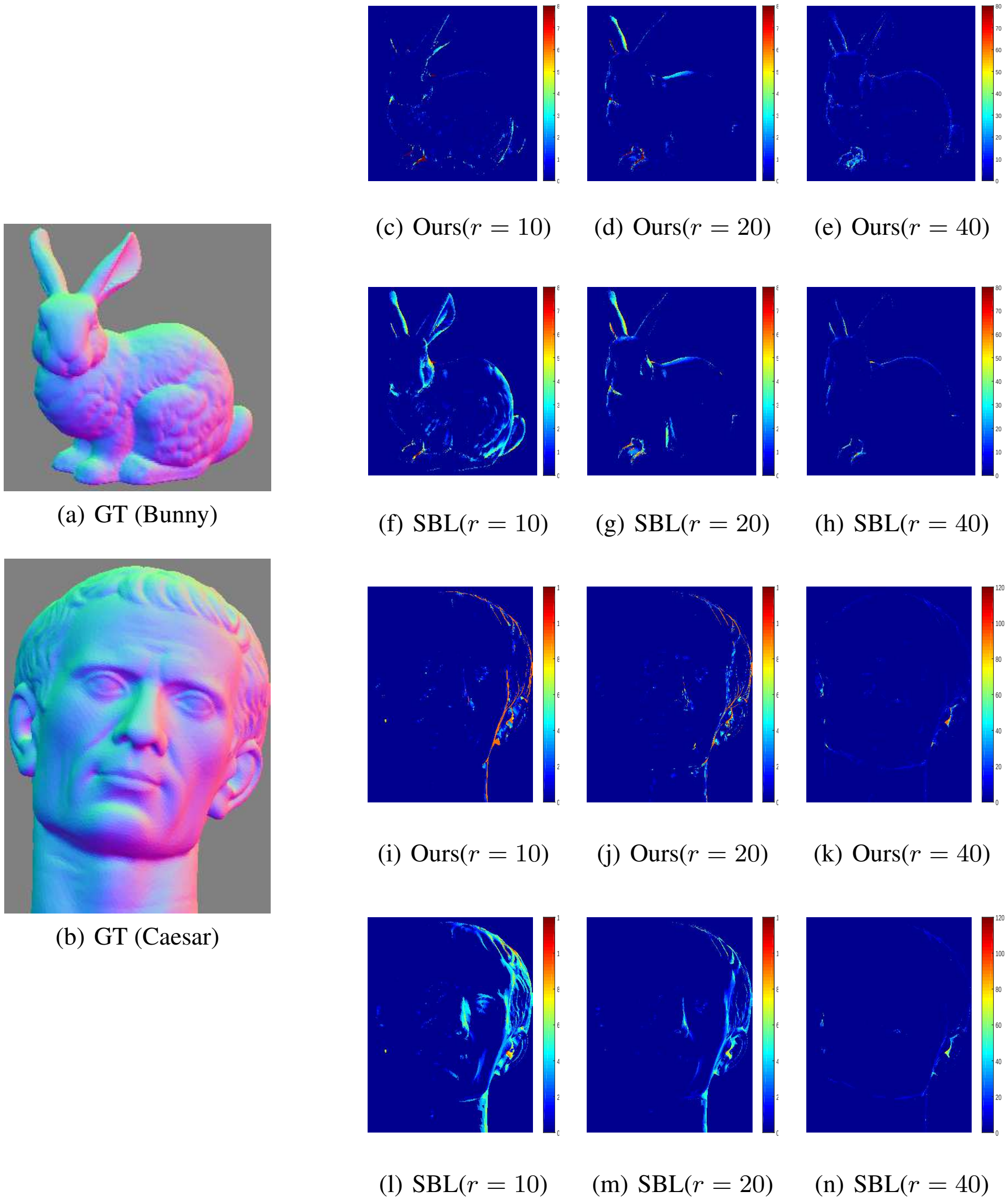

(a) GT (Bunny)

(b) GT (Caesar)

(c) Ours($r = 10$) (d) Ours($r = 20$) (e) Ours($r = 40$)

(f) SBL($r = 10$) (g) SBL($r = 20$) (h) SBL($r = 40$)

(i) Ours($r = 10$) (j) Ours($r = 20$) (k) Ours($r = 40$)

(l) SBL($r = 10$) (m) SBL($r = 20$) (n) SBL($r = 40$)

Figure 7: Photometric stereo reconstruction error maps with different numbers ($r$) of gray-scale images. These correspond with Table (1) results.

Table 3: Results of models with different capacities. There are three main capacity control factors: number of hidden units, number of LSTM stacked layers, and number of LSTM unrolling steps. For various capacities, results are listed for different total number of parameters, training time per epoch(sec.), and strict-accuracy result are listed.

| #Hidden | #Layers | #Unroll | #Parameters | Time | S-Acc |
|---|---|---|---|---|---|
| 200 | 2 | 5 | 1089300 | 99.333 | 0.0524 |
| 200 | 2 | 8 | 1149300 | 131.723 | 0.2211 |
| 200 | 2 | 11 | 1209300 | 172.013 | 0.4691 |
| 200 | 2 | 14 | 1269300 | 206.084 | 0.4707 |
| 200 | 2 | 17 | 1329300 | 240.847 | 0.5060 |
| 200 | 3 | 5 | 2497700 | 160.404 | 0.1455 |
| 200 | 3 | 8 | 2557700 | 234.113 | 0.4146 |
| 200 | 3 | 11 | 2617700 | 309.859 | 0.4776 |
| 200 | 3 | 14 | 2677700 | 383.898 | 0.5319 |
| 200 | 3 | 17 | 2737700 | 446.197 | 0.6011 |
| 200 | 4 | 5 | 4787300 | 254.694 | 0.2561 |
| 200 | 4 | 8 | 4847300 | 380.841 | 0.5619 |
| 200 | 4 | 11 | 4907300 | 517.010 | 0.5802 |
| 200 | 4 | 14 | 4967300 | 631.771 | 0.6046 |
| 200 | 4 | 17 | 5027300 | 764.149 | 0.6156 |
| 425 | 2 | 5 | 4609650 | 173.272 | 0.0976 |
| 425 | 2 | 8 | 4737150 | 235.728 | 0.4334 |
| 425 | 2 | 11 | 4864650 | 312.642 | 0.6087 |
| 425 | 2 | 14 | 4992150 | 378.839 | 0.6595 |
| 425 | 2 | 17 | 5119650 | 442.985 | 0.6697 |
| 425 | 3 | 5 | 10949375 | 316.927 | 0.2598 |
| 425 | 3 | 8 | 11076875 | 470.105 | 0.6043 |
| 425 | 3 | 11 | 11204375 | 616.227 | 0.6584 |
| 425 | 3 | 14 | 11331875 | 763.941 | 0.6359 |
| 425 | 3 | 17 | 11459375 | 927.643 | 0.6427 |
| 425 | 4 | 5 | 21265400 | 540.034 | 0.3653 |
| 425 | 4 | 8 | 21392900 | 821.973 | 0.6376 |
| 425 | 4 | 11 | 21520400 | 1096.844 | 0.6372 |
| 425 | 4 | 14 | 21647900 | 1389.964 | 0.6569 |
| 425 | 4 | 17 | 21775400 | 1655.417 | 0.6618 |
| 600 | 2 | 11 | 9387700 | 461.414 | 0.6494 |
| 600 | 2 | 14 | 9567700 | 568.918 | 0.6795 |
| 600 | 2 | 17 | 9747700 | 698.721 | 0.6682 |

# 11 Appendix E: Technical Proofs

Here we present proofs of our technical propositions.

**Proof of Proposition 1**

It has been demonstrated in [41] that using

$$w_i^{(t+1)} = \left[\boldsymbol{\phi}_i^\top \left(\lambda \boldsymbol{I} + \boldsymbol{\Phi}\boldsymbol{\Gamma}^{(t)}\boldsymbol{\Phi}^\top\right)^{-1} \boldsymbol{\phi}_i\right]^{\frac{1}{2}} \tag{35}$$

will satisfy the stated conditions of Proposition 1. Now assume that $\boldsymbol{\Gamma}^{(t)}$ is full rank or invertible, i.e., $\gamma_j^{(t)} > 0$ for all $j$. Using the matrix inversion lemma, we have

$$\boldsymbol{\phi}_i^\top \left(\lambda \boldsymbol{I} + \boldsymbol{\Phi}\boldsymbol{\Gamma}^{(t)}\boldsymbol{\Phi}^\top\right)^{-1} \boldsymbol{\phi}_i = \tfrac{1}{\lambda}\boldsymbol{\phi}_i^\top \left(\boldsymbol{I} - \tfrac{1}{\lambda}\boldsymbol{\Phi}\left[\left(\boldsymbol{\Gamma}^{(t)}\right)^{-1} + \tfrac{1}{\lambda}\boldsymbol{\Phi}^\top\boldsymbol{\Phi}\right]^{-1}\boldsymbol{\Phi}^\top\right)\boldsymbol{\phi}_i. \tag{36}$$

Given that the matrix inverse is a convex function, and that additive translations preserve convexity, it follows that $\frac{1}{\lambda^2}\boldsymbol{\phi}_i^\top \boldsymbol{\Phi} \left[ \left( \boldsymbol{\Gamma}^{(t)} \right)^{-1} + \frac{1}{\lambda} \boldsymbol{\Phi}^\top \boldsymbol{\Phi} \right]^{-1} \boldsymbol{\Phi}^\top \boldsymbol{\phi}_i$ is a convex function of $\left( \boldsymbol{\Gamma}^{(t)} \right)^{-1}$. Therefore the negation of this term is concave, and so overall (36) is a concave function of $\left( \boldsymbol{\Gamma}^{(t)} \right)^{-1}$. This then implies that we can express (36) as a minimization of upper-bounding hyperplanes via

$$\boldsymbol{\phi}_i^\top \left( \lambda \boldsymbol{I} + \boldsymbol{\Phi}\boldsymbol{\Gamma}^{(t)}\boldsymbol{\Phi}^\top \right)^{-1} \boldsymbol{\phi}_i \;=\; \min_{\boldsymbol{z}} g(\boldsymbol{z}) + \sum_{j=1}^{m} \frac{f(z_j)}{\gamma_i} \tag{37}$$

for some functions $f$ and $g$ and variational parameters $\boldsymbol{z} = [z_1, \ldots, z_m]^\top$. Such a decomposition is not unique; however, using linear algebraic manipulations, it can be easily verified that

$$\boldsymbol{\phi}_i^\top \left( \lambda \boldsymbol{I} + \boldsymbol{\Phi}\boldsymbol{\Gamma}^{(t)}\boldsymbol{\Phi}^\top \right)^{-1} \boldsymbol{\phi}_i \;=\; \min_{\boldsymbol{z}} \frac{1}{\lambda} \|\boldsymbol{\phi}_i - \boldsymbol{\Phi}\boldsymbol{z}\|_2^2 + \sum_{j=1}^{m} \frac{z_j^2}{\gamma_j^{(t)}} \tag{38}$$

is one such viable representation.

To handle the more general case where some $\gamma_j^{(t)} = 0$, we use $\bar{\boldsymbol{\Phi}}$ to denote the columns $\boldsymbol{\phi}_j$ such that $j \in \text{supp}[\gamma]$, and likewise $\bar{\boldsymbol{\Gamma}}^{(t)}$ and $\bar{z}$ the corresponding submatrix of $\boldsymbol{\Gamma}^{(t)}$ and elements of $\boldsymbol{z}$ respectively. It then naturally follows that

$$\begin{aligned}
\boldsymbol{\phi}_i^\top \left( \lambda \boldsymbol{I} + \boldsymbol{\Phi}\boldsymbol{\Gamma}^{(t)}\boldsymbol{\Phi}^\top \right)^{-1} \boldsymbol{\phi}_i &= \boldsymbol{\phi}_i^\top \left( \lambda \boldsymbol{I} + \bar{\boldsymbol{\Phi}}\bar{\boldsymbol{\Gamma}}^{(t)}\bar{\boldsymbol{\Phi}}^\top \right)^{-1} \boldsymbol{\phi}_i \\
&= \min_{\bar{\boldsymbol{z}}} \frac{1}{\lambda} \|\boldsymbol{\phi}_i - \bar{\boldsymbol{\Phi}}\bar{\boldsymbol{z}}\|_2^2 + \sum_{j=1}^{\|\boldsymbol{\gamma}^{(t)}\|_0} \frac{\bar{z}_j^2}{\bar{\gamma}_j^{(t)}} \\
&= \min_{\boldsymbol{z}:\text{supp}[\boldsymbol{z}] \subseteq \text{supp}[\boldsymbol{\gamma}^{(t)}]} \frac{1}{\lambda} \|\boldsymbol{\phi}_i - \boldsymbol{\Phi}\boldsymbol{z}\|_2^2 + \sum_{j \in \text{supp}[\boldsymbol{\gamma}^{(t)}]} \frac{z_j^2}{\gamma_j^{(t)}}.
\end{aligned} \tag{39}$$

**Proof of Proposition 3**

The original SBL objective is given by

$$\mathcal{L}(\gamma) = \boldsymbol{y}^\top \left( \boldsymbol{\Phi}\boldsymbol{\Gamma}\boldsymbol{\Phi}^\top + \lambda \boldsymbol{I} \right)^{-1} \boldsymbol{y} + \log \left| \boldsymbol{\Phi}\boldsymbol{\Gamma}\boldsymbol{\Phi}^\top + \lambda \boldsymbol{I} \right|, \tag{40}$$

where the first term is convex in $\boldsymbol{\gamma}$ while the second is concave, ultimately resulting in a non-convex function. For optimization purposes, it is convenient to decouple elements of $\boldsymbol{\gamma}$ via a series of upper bounds, the iterative minimization of which leads to LSTM-like updates given judicious choices for these bounds.

To begin, we have the linear upper bound

$$h(\gamma) \triangleq \log \left| \boldsymbol{\Phi}\boldsymbol{\Gamma}\boldsymbol{\Phi}^\top + \lambda \boldsymbol{I} \right| \leq h(\widetilde{\gamma}) + (\gamma - \widetilde{\gamma})^\top \nabla h(\widetilde{\gamma}), \tag{41}$$

which is always realizable for any $\widetilde{\gamma} \in \mathbb{R}_+^m$ given the concavity of $h(\boldsymbol{\gamma})$ [10]. This bound decouples individual elements of $\boldsymbol{\gamma}$ into a linear summation that facilitates convenient, separable optimization. Analogously, for the data-dependent term we have

$$\boldsymbol{y}^\top \left( \boldsymbol{\Phi}\boldsymbol{\Gamma}\boldsymbol{\Phi}^\top + \lambda \boldsymbol{I} \right)^{-1} \boldsymbol{y} \leq \tfrac{1}{\lambda} \|\boldsymbol{y} - \boldsymbol{\Phi}\boldsymbol{u}\|_2^2 + \boldsymbol{u}^\top \boldsymbol{\Gamma}^{-1} \boldsymbol{u}. \tag{42}$$

This bound holds for all $\boldsymbol{u} \in \mathbb{R}^m$, with equality when $\boldsymbol{u} = \boldsymbol{\Gamma}\boldsymbol{\Phi}^\top \left( \lambda \boldsymbol{I} + \boldsymbol{\Phi}\boldsymbol{\Gamma}\boldsymbol{\Phi}^\top \right)^{-1} \boldsymbol{y}$ [42].[7] Although the r.h.s. of (42) has effectively decoupled $\boldsymbol{\gamma}$ (given that $\boldsymbol{\Gamma}$ is diagonal, $\boldsymbol{u}^\top \boldsymbol{\Gamma}^{-1} \boldsymbol{u}$ is separable), it has introduced new auxiliary variables $\boldsymbol{u}$ which are inter-mixed via a $\boldsymbol{\Phi}$-dependent norm. However, we can further bound this term using

$$f(\boldsymbol{u}) \triangleq \tfrac{1}{\lambda}\|\boldsymbol{y}-\boldsymbol{\Phi}\boldsymbol{u}\|_2^2 \leq f(\widetilde{\boldsymbol{u}}) + (\boldsymbol{u}-\widetilde{\boldsymbol{u}})^\top \nabla f(\widetilde{\boldsymbol{u}}) + \tfrac{1}{2\mu}\|\boldsymbol{u}-\widetilde{\boldsymbol{u}}\|_2^2, \tag{43}$$

for any $\widetilde{\boldsymbol{u}} \in \mathbb{R}^m$ provided that $\mu \in \left(0, \lambda/\left\|\boldsymbol{\Phi}^\top\boldsymbol{\Phi}\right\|\right]$. This occurs because $\nabla f(\boldsymbol{u})$ is Lipshitz continuous with Lipschitz constant $\frac{1}{\lambda}\left\|\boldsymbol{\Phi}^\top\boldsymbol{\Phi}\right\|$, in which case a quadratic upper bound can always be constructed as in (43).

[7]Additionally, if some $\boldsymbol{\gamma}_j = 0$ while $u_j \neq 0$, we simply define this bound to be infinity. All subsequent update rules are well-defined regardless.

Combining terms, we arrive at the auxiliary objective function

$$\mathcal{L}(\boldsymbol{\gamma},\widetilde{\boldsymbol{\gamma}},\boldsymbol{u},\widetilde{\boldsymbol{u}}) \triangleq h(\widetilde{\boldsymbol{\gamma}}) + (\boldsymbol{\gamma}-\widetilde{\boldsymbol{\gamma}})^\top \nabla h(\widetilde{\boldsymbol{\gamma}}) + \boldsymbol{u}^\top\boldsymbol{\Gamma}^{-1}\boldsymbol{u} + f(\widetilde{\boldsymbol{u}}) + (\boldsymbol{u}-\widetilde{\boldsymbol{u}})^\top \nabla f(\widetilde{\boldsymbol{u}}) + \tfrac{1}{2\mu}\|\boldsymbol{u}-\widetilde{\boldsymbol{u}}\|_2^2, \tag{44}$$

where $\widetilde{\boldsymbol{\gamma}}$, $\boldsymbol{u}$, and $\widetilde{\boldsymbol{u}}$ can be viewed in this context as additional latent variables, sometimes referred to as variational paramters. And by design, for any $\boldsymbol{\gamma}$ we have that

$$\mathcal{L}(\boldsymbol{\gamma}) \;=\; \min_{\widetilde{\boldsymbol{\gamma}},\boldsymbol{u},\widetilde{\boldsymbol{u}}} \mathcal{L}(\boldsymbol{\gamma},\widetilde{\boldsymbol{\gamma}},\boldsymbol{u},\widetilde{\boldsymbol{u}}) \;\leq\; \mathcal{L}(\boldsymbol{\gamma},\widetilde{\boldsymbol{\gamma}},\boldsymbol{u},\widetilde{\boldsymbol{u}}). \tag{45}$$

Additionally, given that this minimization can be accomplished exactly using the stated updates from Section 2. The details are as follows.

Assume that we would like to reduce $\mathcal{L}(\boldsymbol{\gamma})$ starting from some arbitrary point $\boldsymbol{\gamma}^{(t)}$. If we choose

$$\widetilde{\boldsymbol{\gamma}}^{(t)} = \boldsymbol{\gamma}^{(t)}, \qquad \boldsymbol{u}^{(t)} = \boldsymbol{\Gamma}^{(t)}\boldsymbol{\Phi}^\top\left(\lambda\boldsymbol{I}+\boldsymbol{\Phi}\boldsymbol{\Gamma}^{(t)}\boldsymbol{\Phi}^\top\right)^{-1}\boldsymbol{y}, \qquad \widetilde{\boldsymbol{u}}^{(t)} = \boldsymbol{u}^{(t)}, \tag{46}$$

then $\mathcal{L}\left(\boldsymbol{\gamma}^{(t)}\right) = \mathcal{L}\left(\boldsymbol{\gamma}^{(t)},\widetilde{\boldsymbol{\gamma}}^{(t)},\boldsymbol{u}^{(t)},\widetilde{\boldsymbol{u}}^{(t)}\right)$ by construction, i.e., these values simultaneously optimize $\mathcal{L}(\boldsymbol{\gamma},\widetilde{\boldsymbol{\gamma}},\boldsymbol{u},\widetilde{\boldsymbol{u}}) \;\leq\; \mathcal{L}(\boldsymbol{\gamma},\widetilde{\boldsymbol{\gamma}},\boldsymbol{u},\widetilde{\boldsymbol{u}})$ per our structuring of the respective bounds. Our strategy will now be to solve

$$\min_{\boldsymbol{\gamma},\boldsymbol{u}} \mathcal{L}\left(\boldsymbol{\gamma},\widetilde{\boldsymbol{\gamma}}^{(t)},\boldsymbol{u},\widetilde{\boldsymbol{u}}^{(t)}\right) \tag{47}$$

in closed form in order to obtain a new $\boldsymbol{\gamma}^{(t+1)}$ that reduces the original objective function $\mathcal{L}(\boldsymbol{\gamma})$. For this purpose we define $\boldsymbol{w}^{(t)}$ such that

$$\left(\boldsymbol{w}^{(t)}\right)^2 = \nabla h\left(\widetilde{\boldsymbol{\gamma}}^{(t)}\right) = \text{diag}\left[\boldsymbol{\Phi}^\top\left(\lambda\boldsymbol{I}+\boldsymbol{\Phi}\boldsymbol{\Gamma}^{(t)}\boldsymbol{\Phi}^\top\right)^{-1}\boldsymbol{\Phi}\right], \tag{48}$$

where the squaring operator is applied element-wise and the gradient is calculated using standard formulae. Note that this representation is always possible given that $\nabla h(\widetilde{\boldsymbol{\gamma}})$ must always have non-negative elements since $h$ is a non-decreasing, concave function.

By excluding irrelevant terms, taking derivatives, and equating to zero, it follows that

$$\left(\boldsymbol{w}^{(t)}\right)^{-1}\odot|\boldsymbol{u}| \;=\; \arg\min_{\boldsymbol{\gamma}} \mathcal{L}\left(\boldsymbol{\gamma},\widetilde{\boldsymbol{\gamma}}^{(t)},\boldsymbol{u},\widetilde{\boldsymbol{u}}^{(t)}\right) \;\equiv\; \arg\min_{\boldsymbol{\gamma}} \sum_i \left[\left(w_i^{(t)}\right)^2\gamma_i^{(t)} + \frac{u_i^2}{\gamma_i^{(t)}}\right]. \tag{49}$$

Plugging this value into the $\boldsymbol{\gamma}$-dependent terms from $\mathcal{L}\left(\boldsymbol{\gamma},\widetilde{\boldsymbol{\gamma}}^{(t)},\boldsymbol{u},\widetilde{\boldsymbol{u}}^{(t)}\right)$, we find that

$$\boldsymbol{\gamma}^\top\nabla h(\widetilde{\boldsymbol{\gamma}}) + \boldsymbol{u}^\top\boldsymbol{\Gamma}^{-1}\boldsymbol{u} \;\equiv\; 2\boldsymbol{w}^{(t)}\odot|\boldsymbol{u}|. \tag{50}$$

Therefore, a conditionally optimal version of $\boldsymbol{u}$ can be achieved by solving

$$\begin{aligned}
(\boldsymbol{u}^*)^{(t)} &\triangleq \arg\min_{\boldsymbol{u}}\; \mathcal{L}\left(\left[\boldsymbol{w}^{(t)}\right]^{-1}\odot|\boldsymbol{u}|,\widetilde{\boldsymbol{\gamma}}^{(t)},\boldsymbol{u},\widetilde{\boldsymbol{u}}^{(t)}\right)\\
&\equiv \arg\min_{\boldsymbol{u}}\; 2\boldsymbol{w}^{(t)}\odot|\boldsymbol{u}| + \boldsymbol{u}^\top\nabla f\left(\widetilde{\boldsymbol{u}}^{(t)}\right) + \tfrac{1}{2\mu}\left\|\boldsymbol{u}-\widetilde{\boldsymbol{u}}^{(t)}\right\|_2^2\\
&\equiv \arg\min_{\boldsymbol{u}}\; 2\boldsymbol{w}^{(t)}\odot|\boldsymbol{u}| + \tfrac{1}{2\mu}\left\|\boldsymbol{u}-\left[\widetilde{\boldsymbol{u}}^{(t)}-\mu\nabla f\left(\widetilde{\boldsymbol{u}}^{(t)}\right)\right]\right\|_2^2.
\end{aligned} \tag{51}$$

This expression can be optimized independently across each $u_i$, leading to

$$\begin{aligned}
(u_i^*)^{(t)} &= S_{2\lambda w_i^{(t)}}\left(\widetilde{u}_i^{(t)} - \mu\left[\nabla f\left(\widetilde{\boldsymbol{u}}^{(t)}\right)\right]_i\right)\\
&= S_{2\lambda w_i^{(t)}}\left(\widetilde{u}_i^{(t)} + \mu\left[\boldsymbol{\Phi}^\top\left(\boldsymbol{y}-\boldsymbol{\Phi}\widetilde{\boldsymbol{u}}^{(t)}\right)\right]_i\right)
\end{aligned} \tag{52}$$

where $S_\omega$ is a soft threshold operator. Moreover, based on (49), it follows that

$$\boldsymbol{\gamma}^{(t+1)} = \left(\boldsymbol{w}^{(t)}\right)^{-1} \odot \left|(\boldsymbol{u}^*)^{(t)}\right| \tag{53}$$

will be such that

$$\mathcal{L}\left(\boldsymbol{\gamma}^{(t+1)}\right) \leq \mathcal{L}\left(\boldsymbol{\gamma}^{(t+1)}, \widetilde{\boldsymbol{\gamma}}^{(t)}, (\boldsymbol{u}^*)^{(t)}, \widetilde{\boldsymbol{u}}^{(t)}\right) \leq \mathcal{L}\left(\boldsymbol{\gamma}^{(t)}, \widetilde{\boldsymbol{\gamma}}^{(t)}, \boldsymbol{u}^{(t)}, \widetilde{\boldsymbol{u}}^{(t)}\right) = \mathcal{L}\left(\boldsymbol{\gamma}^{(t)}\right). \tag{54}$$

Therefore, by following the above process, $\mathcal{L}(\boldsymbol{\gamma})$ will be reduced (or left unchanged). One attractive feature of this formulation is that $\boldsymbol{\gamma}$ can be optimized jointly with at least one set of variational parameters (in this case $\boldsymbol{u}$), as opposed to most majorization-minimization strategies [25] that fix the upper bound before minimizing the original variables (in this case $\boldsymbol{\gamma}$).

If we choose $\boldsymbol{\alpha}(\boldsymbol{\gamma}) = \mathbf{1}$ and $\boldsymbol{\beta}(\boldsymbol{\gamma}) = \mathbf{0}$, then these steps exactly mirror the revised SBL iterations from Section 2 once we define $\boldsymbol{x}^{(t+1)} \triangleq (\boldsymbol{u}^*)^{(t)}$ and note that $\boldsymbol{\sigma}_{in}^{(t+1)} \odot \bar{\boldsymbol{x}}^{(t+1)}$ is tantamount to soft-thresholding. The more general case follows with a few additional manipulations.

Following the updates described above, we have

$$\begin{aligned}
\mathcal{L}\left(\boldsymbol{\gamma}^{(t)}\right) &= \mathcal{L}\left(\boldsymbol{\gamma}^{(t)}, \widetilde{\boldsymbol{\gamma}}^{(t)}, \boldsymbol{u}^{(t)}, \widetilde{\boldsymbol{u}}^{(t)}\right) \\
&= h\left(\widetilde{\boldsymbol{\gamma}}^{(t)}\right) + \left(\boldsymbol{u}^{(t)}\right)^\top \left(\boldsymbol{\Gamma}^{(t)}\right)^{-1} \boldsymbol{u}^{(t)} + \tfrac{1}{\lambda}\left\|\boldsymbol{y} - \boldsymbol{\Phi}\boldsymbol{u}^{(t)}\right\|_2^2 \\
&\geq h\left(\widetilde{\boldsymbol{\gamma}}^{(t)}\right) - \left(\widetilde{\boldsymbol{\gamma}}^{(t)}\right)^\top \nabla h\left(\widetilde{\boldsymbol{\gamma}}^{(t)}\right) + 2\boldsymbol{w}^{(t)} \odot \left|\boldsymbol{u}^{(t)}\right| + \tfrac{1}{\lambda}\left\|\boldsymbol{y} - \boldsymbol{\Phi}\boldsymbol{u}^{(t)}\right\|_2^2 \\
&= h\left(\widetilde{\boldsymbol{\gamma}}^{(t)}\right) - \left(\widetilde{\boldsymbol{\gamma}}^{(t)}\right)^\top \nabla h\left(\widetilde{\boldsymbol{\gamma}}^{(t)}\right) + 2\boldsymbol{w}^{(t)} \odot \left|\boldsymbol{u}^{(t)}\right| + f\left(\widetilde{\boldsymbol{u}}^{(t)}\right) \\
&\quad + \left(\boldsymbol{u}^{(t)} - \widetilde{\boldsymbol{u}}^{(t)}\right)^\top \nabla f\left(\widetilde{\boldsymbol{u}}^{(t)}\right) + \tfrac{1}{2\mu}\left\|\boldsymbol{u}^{(t)} - \widetilde{\boldsymbol{u}}^{(t)}\right\|_2^2
\end{aligned} \tag{55}$$

given that presently $\boldsymbol{u}^{(t)} = \widetilde{\boldsymbol{u}}^{(t)}$. Previously we optimized this expression with respect to $\boldsymbol{u}$ and obtained the soft-threshold estimator $(\boldsymbol{u}^*)^{(t)}$. However, suppose we instead evaluate at an alternative point $(\boldsymbol{u}')^{(t)}$ defined recursively such that

$$(\boldsymbol{u}')^{(t)} \equiv \boldsymbol{x}^{(t+1)} = \boldsymbol{\beta}\left(\boldsymbol{\gamma}^{(t)}\right) \odot \boldsymbol{x}^{(t)} + \boldsymbol{\alpha}\left(\boldsymbol{\gamma}^{(t)}\right) \odot (\boldsymbol{u}^*)^{(t)}. \tag{56}$$

Then finally we have

$$\begin{aligned}
\mathcal{L}\left(\boldsymbol{\gamma}^{(t)}\right) &\geq h\left(\widetilde{\boldsymbol{\gamma}}^{(t)}\right) - \left(\widetilde{\boldsymbol{\gamma}}^{(t)}\right)^\top \nabla h\left(\widetilde{\boldsymbol{\gamma}}^{(t)}\right) + 2\boldsymbol{w}^{(t)} \odot \left|\boldsymbol{u}^{(t)}\right| + \tfrac{1}{\lambda}\left\|\boldsymbol{y} - \boldsymbol{\Phi}\boldsymbol{u}^{(t)}\right\|_2^2 \\
&\geq h\left(\widetilde{\boldsymbol{\gamma}}^{(t)}\right) - \left(\widetilde{\boldsymbol{\gamma}}^{(t)}\right)^\top \nabla h\left(\widetilde{\boldsymbol{\gamma}}^{(t)}\right) + 2\boldsymbol{w}^{(t)} \odot \left|(\boldsymbol{u}')^{(t)}\right| + \tfrac{1}{\lambda}\left\|\boldsymbol{y} - \boldsymbol{\Phi}\,(\boldsymbol{u}')^{(t)}\right\|_2^2 \\
&= h\left(\widetilde{\boldsymbol{\gamma}}^{(t)}\right) + \left(\boldsymbol{\gamma}^{(t+1)} - \widetilde{\boldsymbol{\gamma}}^{(t)}\right)^\top \nabla h\left(\widetilde{\boldsymbol{\gamma}}^{(t)}\right) + \left(\boldsymbol{x}^{(t+1)}\right)^\top \left(\boldsymbol{\Gamma}^{(t+1)}\right)^{-1} \boldsymbol{x}^{(t+1)} \\
&\quad + \tfrac{1}{\lambda}\left\|\boldsymbol{y} - \boldsymbol{\Phi}\boldsymbol{x}^{(t+1)}\right\|_2^2 \\
&\geq \mathcal{L}\left(\boldsymbol{\gamma}^{(t+1)}\right),
\end{aligned} \tag{57}$$

where now $\boldsymbol{\gamma}^{(t+1)} = \left(\boldsymbol{w}^{(t)}\right)^{-1} \odot \left|(\boldsymbol{u}')^{(t)}\right|$. The first inequality follows from (55), the second from the monotone cell update property, and the third via the original construction of the majorization-minimization algorithm. This process then exactly mirrors the iterations from Section 2, with guaranteed cost function descent.